\documentclass[sigconf]{acmart}
\AtBeginDocument{%
  \providecommand\BibTeX{{%
    \normalfont B\kern-0.5em{\scshape i\kern-0.25em b}\kern-0.8em\TeX}}}

\setcopyright{acmcopyright}
\copyrightyear{2025}
\acmYear{2025}
\setcopyright{acmlicensed}\acmConference[ICMR '25]{Proceedings of the 2025 International Conference on Multimedia Retrieval}{June 30-July 3, 2025}{Chicago, IL, USA}
\acmBooktitle{Proceedings of the 2025 International Conference on Multimedia Retrieval (ICMR '25), June 30-July 3, 2025, Chicago, IL, USA}
\acmDOI{10.1145/3731715.3733291}
\acmISBN{979-8-4007-1877-9/2025/06}

\usepackage{algorithm}
\usepackage{algorithmic}
\usepackage{subfigure}
\usepackage{multirow}
\usepackage[shortlabels]{enumitem}
\newcommand{\stitle}[1]{\vspace{1ex} \noindent{\bf #1}}
\usepackage{bbding}
\usepackage[inkscapelatex=false]{svg}
\title{CoATA: Effective Co-Augmentation of Topology and Attribute for Graph Neural Networks}

	\author{Tao Liu}
	\affiliation{%
		\institution{College of Computer and Information Science, Southwest University}
              \city{Chongqing}
  \country{China}
	}
	\email{taoliu.swu@gmail.com}

	\author{Longlong Lin}
\affiliation{%
	\institution{College of Computer and Information Science, Southwest University}
          \city{Chongqing}
  \country{China}
}
\email{longlonglin@swu.edu.cn}

	\author{Yunfeng Yu}
\affiliation{%
	\institution{College of Computer and Information Science, Southwest University}
          \city{Chongqing}
  \country{China}
}
\email{YunfengYu817@outlook.com}

	\author{Xi Ou}
\affiliation{%
	\institution{College of Computer and Information Science, Southwest University}
          \city{Chongqing}
  \country{China}
}
\email{xiou.cs@outlook.com}

	\author{Youan Zhang}
\affiliation{%
	\institution{College of Computer and Information Science, Southwest University}
          \city{Chongqing}
  \country{China}
}
\email{youan0529@email.swu.edu.cn}

	\author{Zhiqiu Ye}
\affiliation{%
	\institution{College of Computer and Information Science, Southwest University}
          \city{Chongqing}
  \country{China}
}
\email{zqiuye@outlook.com}
	\author{Tao Jia}
\affiliation{%
	\institution{College of Computer and Information Science, Southwest University}
          \city{Chongqing}
  \country{China}
}
\email{tjia@swu.edu.cn}
\renewcommand{\shortauthors}{Tao Liu et al.}

\begin{document}

%\maketitle
%one-sided变为 single-dimensional
%, which adeptly explore both topology structures and node attributes through aggregation and transformation mechanisms within neighborhoods

\begin{abstract}
Graph Neural Networks (GNNs) have garnered substantial attention due to their remarkable capability in learning graph representations. However,  real-world graphs often exhibit substantial noise and incompleteness, which severely degrades the
performance of GNNs. Existing methods typically address this issue through single-dimensional augmentation, focusing either on refining topology structures or perturbing node attributes, thereby overlooking the deeper interplays between the two. To bridge this gap, this paper presents CoATA, a dual-channel GNN framework specifically designed for the Co-Augmentation of Topology and Attribute. Specifically, CoATA first propagates structural signals to enrich and denoise node attributes. Then, it projects the enhanced attribute space into a node-attribute bipartite graph for further refinement or reconstruction of the underlying structure. Subsequently, CoATA introduces contrastive learning, leveraging prototype alignment and consistency constraints, to facilitate mutual corrections between the augmented and original graphs. Finally, extensive experiments on seven benchmark datasets demonstrate that the proposed CoATA outperforms eleven state-of-the-art baseline methods, showcasing its effectiveness in capturing the synergistic relationship between topology and attributes. 

\end{abstract}




\maketitle

\section{Introduction}
Graphs are foundational data structures for modeling relationships among entities in diverse real-world domains. Graph Neural Networks (GNNs) have emerged as a powerful paradigm to exploit such graph-structured data, driving advances in social media analysis \cite{social1,social2}, recommender systems \cite{lightgcn,DBLP:conf/kdd/YingHCEHL18}, natural language processing natural language processing \cite{DBLP:conf/mir/WangLXL024,DBLP:conf/mir/WangZXLW24}, and computer vision \cite{cvpr,cvpr2}. Classical GNNs (e.g., GAT \cite{gat}, GCN \cite{gcn}, and GraphSAGE \cite{graphsage}) and subsequent refinements \cite{sgc,ppnp,pprgo,gbp} maintain an embedding vector for each node, which is iteratively updated by aggregating the embeddings of its neighbors.

\begin{figure}[t]
\centering
\subfigure[\textbf{Cora}]{\includegraphics[width=0.23\textwidth]{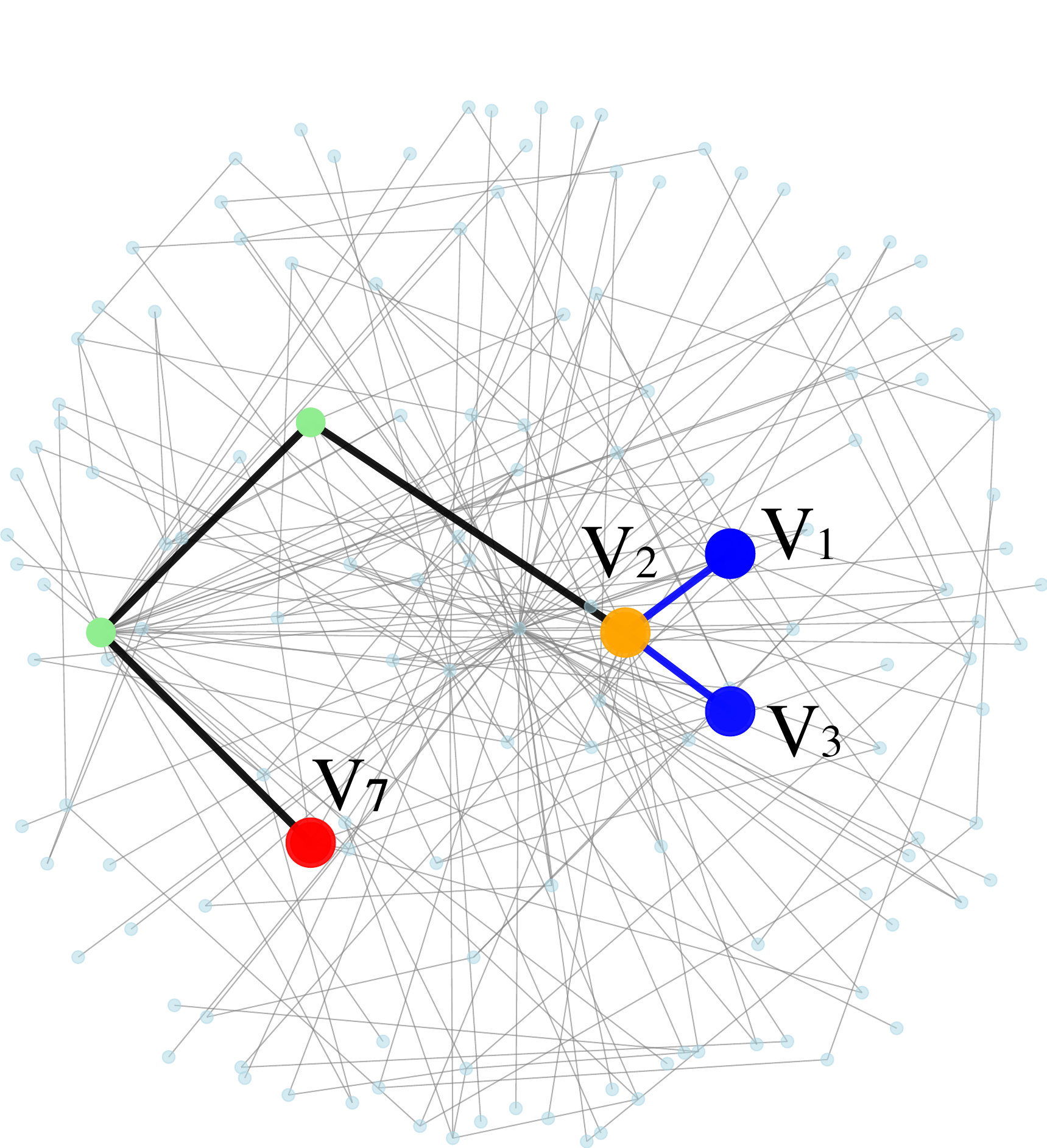}}
%\hspace{4mm}  % 控制两张图片间距
\subfigure[\textbf{Localized View}]{\includegraphics[width=0.23\textwidth]{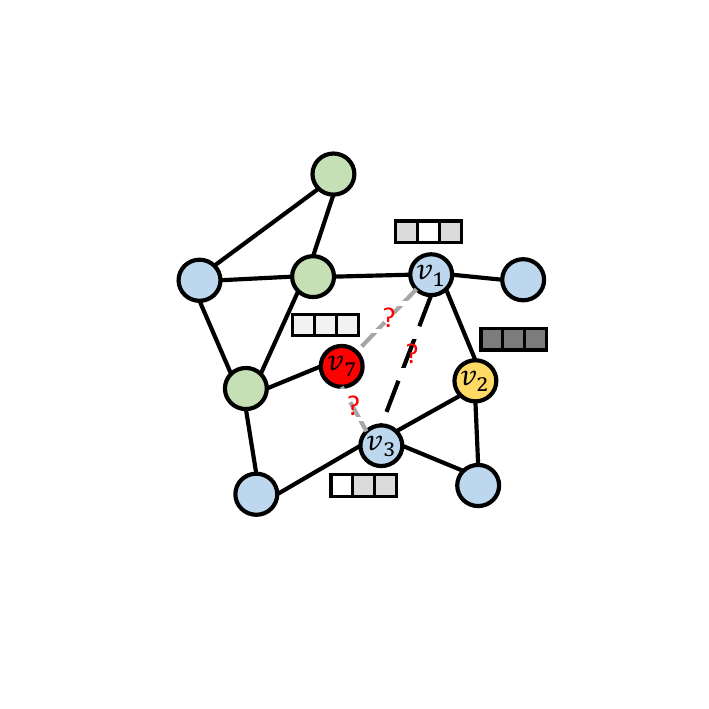}}
\vspace{-5.02mm}
\caption{(a) Visualization of the Cora dataset, where nodes of the same class ($v_2$, $v_7$) are found to be multi-hop away. 
(b) A magnified subgraph where existing graph augmentation methods may fail to establish meaningful connections.}
\vspace{-0.5cm}
\label{fig:intro}
\end{figure}

Despite their success, real-world graphs often suffer from significant noise and incompleteness, with corrupted topology structures and unreliable node attributes,  which can severely degrade the performance of GNNs \cite{assu1,assu2,ddpt}. 
Therefore, numerous graph augmentation methods (Section \ref{sec:existingwork}) have been proposed to alleviate the issues mentioned above. However, existing methods typically focus on only one of these two aspects: \emph{(i)} Topology-oriented methods, such as edge denoising \cite{learndrop}, random walks \cite{dgc}, or auxiliary nodes \cite{addnode}, aim to repair structural gaps but often overlook the frequent attribute corruption. 
 \emph{(ii)} Attribute-oriented methods focus on feature-space enhancements, such as feature (i.e., attribute) masking \cite{DropMessage}, feature propagation \cite{grand}, or feature mixup \cite{gmixup1}. These techniques assume the underlying topology is sufficiently reliable to guide message passing or synthetic feature generation. %2.17：这一段放到下面的例子里面具体说明For example, when two originally highly similar nodes share only a few corrupted attributes,  one-hop cosine similarity may fail to capture their true higher-order relationships, leaving them disconnected.%
Consequently, such \emph{single-dimensional} approaches tend to create a dilemma loop: missing or incorrect edges impede effective feature propagation, while noisy attributes mislead topology refinements, leading to ineffective learning in real-world graphs.

%In the case of \(v_7\), its isolation prevents it from receiving informative signals from the more central nodes \(v_2\).

To illustrate, consider the widely used Cora dataset \cite{cora}, a citation network where nodes represent research publications and edges indicate citation relationships (Fig.\ref{fig:intro}). Due to the sparsity of direct citations or the high-dimensional and noisy nature of textual features, semantically similar papers may remain unlinked. For example, nodes \(v_1\) and \(v_3\) might be from the same cluster due to they share a neighbor \(v_2\). However, another node \(v_7\), despite belonging to the same class with $v_2$, remains disconnected due to a missing direct citation link to \(v_2\) and its multi-hop away from the core cluster. Thus, existing attribute-oriented methods suffer from two critical limitations. First, they only aggregate information from immediate neighbors, meaning that any missing or spurious edges can lead to significant information loss or contamination.  Second, such local propagation mechanisms often lead to over-smoothing when stacking multiple layers,  especially problematic in networks with noisy or heterophilic connections \cite{AdaEdge}. On the other hand,  existing topology-oriented methods often use naive cosine similarity on node attributes to reweight edges also presents challenges. For example, the high-dimensional, sparse nature of the feature vectors means that direct cosine similarity is extremely sensitive to noise; even a few corrupted or missing features can drastically lower the similarity score between nodes that are semantically related in reality. As a consequence, for Fig.\ref{fig:intro}, without effective feature propagation via an intermediary such as \( v_2 \), nodes \( v_1 \) and \( v_3 \) would not exhibit a high cosine similarity based solely on their raw features. Furthermore, since the enriched features of \( v_2 \) fail to propagate to \( v_7 \), the cosine similarity between \( v_7 \) and either \( v_1 \) or \( v_3 \) may collapse to nearly zero, completely masking their latent semantic relevance. Thus, designing the interplay between topology and attributes is both crucial and challenging for achieving robust graph learning.

%

%Addressing these issues requires a framework that integrates both topology and attributes, capturing not only local interactions but also higher-order relationships embedded in the global graph structure. Such a design is crucial for robust graph learning, as it must reconcile and integrate both topology and attributes to reveal genuine underlying affinities.

%However, neither feature propagation nor naive cosine-based reweighting cannot capture the subtle overlaps needed to relate \(v_7\) to \(v_1\) or \(v_3\), thereby perpetuating noise. 

To tackle the challenge above, we propose a novel framework, \underline{Co}-\underline{A}ugmentation of \underline{T}opology and \underline{A}ttribute (\textit{CoATA}), which consists of three principal components: the Topology-Enriched Attributes (\emph{TEA}) module (Section~\ref{sec:topology-enriched}), the Attribute-Informed Topology (\emph{AIT}) module (Section~\ref{sec:attribute-informed}), and Dual-Channel GNN with Prototype Alignment (\emph{DPA}) module (Section~\ref{sec:4.4}). Unlike conventional augmentation techniques that separately enhance topology or attributes, CoATA explicitly co-optimizes both through an iterative correction process. Specifically, while many conventional methods rely solely on the homophily assumption \cite{luansitao}, our TEA module employs higher-order propagation via a residual mechanism to both strengthen local homophilic signals and mitigate the adverse effects of heterophilic or noisy connections. This step can correct or enrich local attributes (e.g., effectively bridging \(v_2\) with \(v_1\) and \(v_3\)) and even propagate informative signals to moderately distant nodes like \(v_7\); However, not all nodes similar to \(v_7\) or those even farther from the central hub can reliably receive the propagated information, as the reach of TEA remains limited. Hence, AIT constructs a node-attribute bipartite graph and leverages the celebrated proximity metric Personalized PageRank \cite{DBLP:ppr} to traverse multi-hop feature pathways on a global scale, forging new or refined edges (e.g., eventually connecting these distant nodes to the core cluster once partial overlaps accumulate). As a result, the co-augmented graph better reflects genuine node proximities than raw structural or naive attribute-based measures. Finally, since neither the original nor the augmented graph is perfect, DPA maintains both views in a dual-channel model, which retains raw edges while injecting the co-augmented structure.  To ensure consistent class-level semantics, we devise a prototype-based contrastive loss, guiding both channels to converge on unified and robust representations. This synergy robustly combines local and global signals, mitigating any bias from a single dimension. In a nutshell, we highlight the main contributions as follows:

%A consistency regularization addresses multiple $k$-NN thresholds, and a prototype alignment objective aligns node embeddings class by class. 

\stitle{Co-Augmentation Pipeline.} We present a novel approach that integrates \emph{topology-enriched attributes} and \emph{attribute-informed topology} into a unified framework. This bidirectional process leverages higher-order structural cues to clean attributes and then uses the improved features to correct or supplement the adjacency matrix. 

\stitle{Dual-Channel GNN with Prototype Alignment.} We propose a dual-channel GNN that preserves both the original and the augmented adjacency matrices. We devise a prototype-based contrastive loss to ensure consistent class-level semantics, guiding both channels to converge on unified and robust representations.

\stitle{Extensive Experiments.} We conduct extensive experiments on seven real-world graphs and eleven baselines to validate the effectiveness of our solutions. The results show that the proposed CoATA consistently outperforms most existing augmentation methods, achieving accuracy improvements ranging from 1.1\% to 17.9\%. %For reproducibility, our source codes and datasets are available at https://anonymous.4open.science/r/coata-1922.

\section{Related Work} \label{sec:existingwork} 

\stitle{Topology-oriented Augmentation}
techniques aim to enhance GNNs performance through structural modifications that address incomplete or noisy graph connectivity \cite{ddpt}. Early approaches relied on stochastic edge manipulation, with methods like DropEdge \cite{sturct4:} demonstrating that random edge removal could effectively regularize GNNs and alleviate over-smoothing issues \cite{AdaEdge}. Subsequent research has evolved toward more sophisticated structural adaptation strategies, which can be broadly categorized into three paradigms: (i) \textit{Similarity-based methods} leverage node similarity metrics to guide structural modifications. Works such as \cite{learnsturct1,TO-GCN,IDGL,learndrop,dgc} compute pairwise node similarities to dynamically adjust edge connections, thereby better capturing latent relational patterns through adaptive topology refinement. (ii) \textit{Graph regeneration approaches} employ generative models to reconstruct graph structures. Techniques including \cite{learnsturct3,learnsturct2,GTS,GEN,MH-Aug} utilize graph autoencoders \cite{GAE} and stochastic block models \cite{SBM} to learn edge formation dynamics, generating synthetic topologies that preserve critical structural properties. 
(iii) \textit{Global and high-order augmentation} focus on augmenting graph structure by incorporating global connectivity information or higher-order substructures. For instance, S³-CL \cite{s3cl} employs structure and semantic contrastive learning to enhance global topology awareness. GloGNN \cite{glognn} utilizes the global adjacency matrix to transcend local neighborhood limitations.  PSA-GNN \cite{psagnn} introduces high-order subgraphs (e.g., motifs, cliques), thus extending standard node-level connectivity to richer topological constructs.

%However, Simp-GCN \cite{simpgnn} reveals fundamental limitations of edge-level modifications: when original connectivity patterns contain systematic biases (e.g., missing high-order relationships or misleading local adjacencies), superficial edge adjustments fail to resolve deeper structural deficiencies. This suggests topology-oriented augmentation remains suboptimal when node attributes contradict structural assumptions.

\stitle{Attribute-oriented Augmentation}
techniques enhance GNNs performance through feature-space modifications, primarily employing three strategies: feature masking, feature propagation, and feature mixup. Feature masking \cite{GRACE,bgrl} randomly obscures feature dimensions during training to prevent over-reliance on specific attributes. For example, DropMessage \cite{DropMessage} extends traditional dropout by probabilistically masking message matrix elements, enabling fine-grained regularization. Feature propagation \cite{GRAND+,s3cl,ddpt} iteratively aggregates neighborhood features to construct noise-resistant representations that capture structural context.  For example, GRAND \cite{grand} leverages multiple feature propagation to create diverse augmented graphs, exposing models to varied structural contexts for improved robustness. SimP-GCN \cite{simpgnn} focuses on preserving original feature similarity to prevent distortions during message passing. By constraining the extent of information propagation, SimP-GCN mitigates over-smoothing and retains critical attribute-level distinctions. Feature mixup \cite{gmixup1,gmixup2,graphmix} generates synthetic samples through convex feature combinations, improving model generalizability. For example, GeoMix \cite{geomix}  integrates mixup operations within propagation steps, generating interpolated features that preserve topological smoothness.

\section{Preliminaries} \label{sec:pro}
\stitle{Basic Notations.} Given an undirected and unweighted graph $G = (V, E)$, where $V$ represents the node set with $n=|V|$ nodes, and $E$ denotes the edge set with $m=|E|$ edges. We define the adjacency matrix $\mathbf{A} \in \{0, 1\}^{n \times n}$. Namely, $\mathbf{A}_{uv} = 1$ if $(u, v) \in E$; otherwise, $\mathbf{A}_{uv} = 0$. Furthermore, each node $u \in V$ is endowed with an attribute (feature) vector $\mathbf{x}_u \in \mathbb{R}^{k}$, and the collection of these attribute vectors forms the node attribute matrix $\mathbf{X} = [\mathbf{x}_1^\top; \mathbf{x}_2^\top; \ldots; \mathbf{x}_n^\top] \in \mathbb{R}^{n \times k}$, where $k$ signifies the dimensionality of the features. Besides, let $N(u) = \{v \mid (u, v) \in E\}$ be the neighbors of node $u$. We introduce the degree diagonal matrix $\mathbf{D} \in \mathbb{R}^{n \times n}$, which is defined as $\mathbf{D} = \operatorname{diag}(d(u_1), d(u_2), \ldots, d(u_n))$, where $d(u_i) = |N(u_i)|$ represents the degree  of node $u_i$. To facilitate our discussion, we partition the node set $V$ into two subsets: $\mathcal{V}_l$ and $\mathcal{V}_u$, where $\mathcal{V}_l$ contains the labeled nodes, and $\mathcal{V}_u = V \setminus \mathcal{V}_l$ consists of the unlabeled nodes. Our primary goal is to leverage the information provided by the labeled nodes $\mathcal{V}_l$ to predict the labels of the unlabeled nodes $\mathcal{V}_u$.

\stitle{Graph Neural Network (GNN)} is a specialized neural network architecture designed for handling graph-structured data \cite{DBLP:journals/pvldb/MengLLLW24, gcn,graphsage,gat,DBLP:conf/mir/YuLLWOJ24}, which aims to map nodes into low-dimensional embedding vectors, capturing both the topology structures and node attributes. GNN regulates the spread and update of information through defined node aggregation and update functions, allowing nodes to integrate information from their neighboring nodes and refine their own embedding vectors. This iterative process persists until the model convergence: $
h_{u}^{(l+1)} = \text{UPDATE}(h_{u}^{(l)}, \text{AGGREGATE}(h_{v}^{(l)} | v \in N(u)))$,
where $h_{u}^{(l)}$ represents the embedding of node $u$ at layer $l$. The $\text{AGGREGATE}(\cdot)$ function assigns weights to the embeddings of neighboring nodes $v$ relative to node $u$. The $\text{UPDATE}(\cdot)$ function iteratively refines the features of all nodes based on the aggregated information. Among the various GNN models, the Graph Convolutional Network (GCN) \cite{gcn} is ubiquitous, which is stated as  follows:
\begin{equation}\label{formula1}
\mathbf{H}^{(l+1)} = \sigma\left( \mathbf{\tilde{A}} \mathbf{H}^{(l)} \mathbf{W}^{(l)}\right), \quad l = 0, 1, \cdots
\end{equation}
Here, $\mathbf{H}^{(l)}$ denotes the node embedding matrix at layer $l$, with $\mathbf{H}^{(0)} = X$. The propagation matrix $\mathbf{\tilde{A}} = \mathbf{\hat{D}}^{-1/2} \mathbf{\hat{A}} \mathbf{\hat{D}}^{-1/2}$ is derived from the adjacency matrix $\mathbf{A}$, where $\mathbf{\hat{A}} = \mathbf{A} + \mathbf{I}$, $\mathbf{\hat{D}} = \mathbf{D} + \mathbf{I}$, and $\mathbf{I}$ is the identity matrix. $\mathbf{W}^{(l)}$ represents the trainable weight matrix at layer $l$, and $\sigma(\cdot)$ is the activation function (e.g., ReLU).

\stitle{Personalized PageRank (PPR)}  is the cutting-edge proximity metric that measures the relative importance of nodes \cite{DBLP:conf/focs/AndersenCL06,  DBLP:conf/cikm/LinYWWZ0024, DBLP:journals/entropy/LiaoLHL24}. The PPR value ${\Pi}(u,v)$ represents the probability that an $\alpha$-decay random walk starting from node $u$ stops at node $v$. In an $\alpha$-decay random walk, there is an $\alpha$ probability of stopping at the current node or a $(1-\alpha)$ probability of randomly jumping to one of its neighbors. Consequently, the length of an $\alpha$-decay random walk follows a geometric distribution with a success probability of $\alpha$. Therefore, the PPR matrix $\Pi$ can capture the higher-order proximity, which  is formulated as follows:
\begin{equation}\label{formula1}
\Pi = \sum_{r=0}^{\infty} \alpha(1-\alpha)^{r} \cdot \mathbf{P}^{r}
\end{equation}
where $\mathbf{P}$ is the state transition matrix, typically set to $D^{-1}A$.

\section{COATA: The Proposed Solution} \label{sec:our}
%4.1：framework overview  
%4.2：topology-enriched  在这一这里提出同亲假设（引用别人的论文），即同类节点更容易连接在一起，并且指出我们的topology-enriched的增强方式能够获得高阶信息，可以纠正结构上的错误，并且能够使原本不丰富/含有噪声的属性更加准确和丰富，为后续在二部图中的ppr做好铺垫。
%4.3：attribute-informed 沿用4.2节的topology-enriched之后的节点特征矩阵，将其转化为二部图，在此再次强调利用二部图和一些传统的计算特征相似度方法相比的优势。
%4.4：总结4.2，4.3得到共同增强的图之后，沿用原有的公式介绍基于原型对齐的对比学习

%\subsection{Framework Overview}
%\label{subsec:method-overview}
\begin{figure*}[t!]
    \centering
\includegraphics[width=0.8\textwidth]{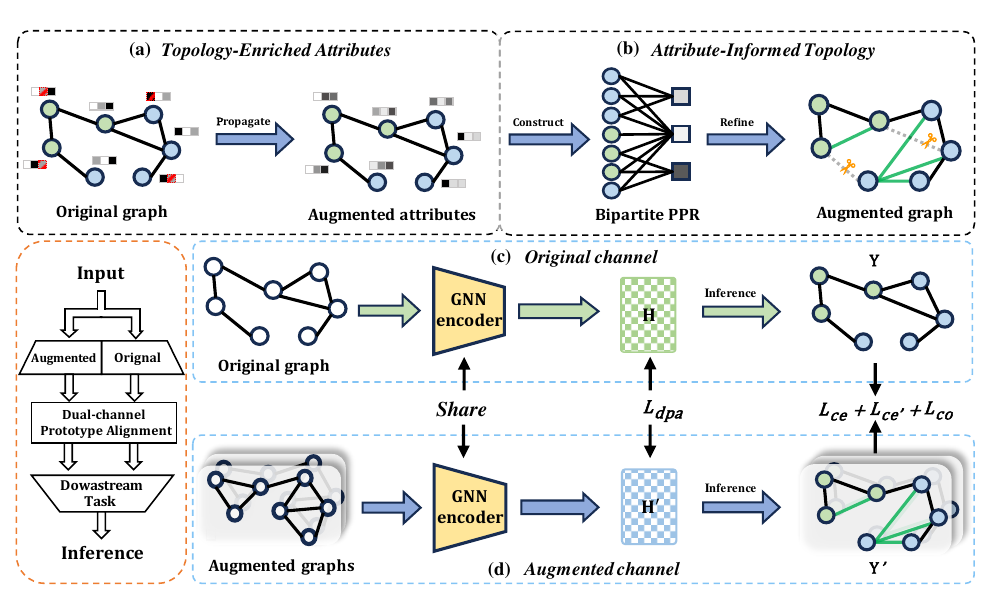}
\caption{The framework of \textbf{CoATA}.  
(a) \textit{TEA} improves node features via \textit{higher-order propagation}; 
(b) \textit{AIT} refines graph structure through a node-attribute bipartite graph.
The model has two channels:  
(c) \textbf{Original channel} processes the raw graph; 
(d) \textbf{Augmented channel} leverages the co-augmented graph.  
Both share a GNN encoder, with \textit{DPA} ensuring cross-view consistency. \textbf{Supervised losses} (\(L_{ce} + L'_{ce}\)) guide label predictions, and \textbf{consistency loss} (\(L_{co}\)) aligns outputs across augmented graphs.}
    \label{fig:framwork}
\end{figure*}

We propose a dual-channel approach \textit{CoATA} for \underline{Co}-\underline{A}ugmentation of \underline{T}opology and \underline{A}ttribute(Fig.~\ref{fig:framwork}), Specifically, \textit{CoATA} first utilizes the Topology-Enriched Attribute  (\textit{TEA})  module to incorporate higher-order structural signals, thereby refining and denoising node features (Section ~\ref{sec:topology-enriched}). Next, the enriched features construct a bipartite graph, and the Attribute-Informed Topology  (\textit{AIT}) module applies the PPR metric to this bipartite graph to more effectively capture multi-hop indirect relationships between nodes (Section ~\ref{sec:attribute-informed}). Finally, \textit{CoATA} unifies the raw and augmented graphs
through Dual-Channel GNN with Prototype Alignment (\textit{DPA}), ensuring that node embeddings derived from  the raw and augmented views remain consistent (Section ~\ref{sec:4.4}). %This integrated strategy not only mitigates topology and attribute noise but also leverages complementary signals from each view to achieve improved node representation. 

%\textbf{CoATA} simultaneously maintains: that follows a two-step co-augmentation process:  two channels (i.e., An \textit{original channel} that retains the raw adjacency and node attributes without explicit augmentation, thus serving as a baseline view of the data. An \textit{enhanced channel} )

%4.2  topology-enriched attribute augmentation 
%这里提出同亲假设，并且指出我们的topology-enriched的方式能够获得高阶信息，能够使原本不丰富/含有噪声的属性更加准确和丰富，为后续在二部图中的ppr做好铺垫。 

\subsection{Topology-Enriched Attribute Augmentation}
\label{sec:topology-enriched}

Real-world graphs often exhibit noisy or incomplete connections and may also be \emph{heterophilic}~\cite{luansitao}, meaning that directly connected nodes can belong to different classes. In such cases, relying solely on single-hop neighbors for message passing can lead to feature contamination and insufficient representation. To address this issue, we propose a higher-order augmentation method that leverages multi-hop structural signals. By propagating node attributes over multiple steps, distant but semantically relevant information is incorporated, helping to overcome local inconsistencies or missing edges. This process lays a robust foundation for subsequent stages in our co-augmentation pipeline, especially when the graph structure is imperfect or deviates from traditional homophily assumptions.

\stitle{Higher-order Attribute Propagation.}  
We iteratively propagate the attribute matrix \(\mathbf{X}\) over a normalized adjacency matrix \(\mathbf{\tilde{A}}\). Let \(\mathbf{X}^{(0)} = \mathbf{X}\) denote the initial feature matrix. At each \( l\)-th propagation step, we compute:
\begin{align}\label{eq:propX}
\mathbf{X}^{(l)} = \mathbf{\tilde{A}}\,\mathbf{X}^{(l-1)}.
\end{align}
However, propagating through too many hops may lead to over-mixing of features, especially in heterophilic graphs where local neighborhoods can be misleading. To preserve node-specific information, we incorporate a residual mixing mechanism~\cite{sdg} controlled by the residual coefficient \(\beta \in [0, 1]\):
\begin{align}\label{eq:resX}
\mathbf{X}^{(l)} = (1 - \beta)\,\mathbf{\tilde{A}}\,\mathbf{X}^{(l-1)} + \beta\,\mathbf{X},
\end{align}
%As stated in Theorem~\ref{theo:convergence}, this iterative process converges to a unique fixed point, ensuring the stability of the final node representations.
Algorithm~\ref{alg:augment_fixed} summarizes the procedure of \emph{Topology-Enriched Attribute Augmentation}, demonstrating how to integrate higher-order information while retaining node-specific features.

\begin{algorithm}[t] \footnotesize
\caption{\textit{Topology-Enriched Attribute Augmentation}}
\label{alg:augment_fixed}
\begin{flushleft}

\textbf{Input}: Node feature matrix \(\mathbf{X} \in \mathbb{R}^{n \times k}\), normalized adjacency matrix \(\mathbf{\tilde{A}} \in \mathbb{R}^{n \times n}\), residual coefficient \(\beta \in [0,1]\), propagation steps \(h\).\\
\textbf{Output}: Enriched node feature matrix \(\mathbf{H} \in \mathbb{R}^{n \times k}\).
\end{flushleft}
\begin{algorithmic}[1]
    \STATE \(\mathbf{X}^{(0)} \leftarrow \mathbf{X}\);
    \FOR{\(l=1\) to \(h\)}
        \STATE \(\mathbf{X}' \leftarrow \mathbf{\tilde{A}} \cdot \mathbf{X}^{(l-1)}\) \hfill // Aggregate multi-hop neighbor features
        \STATE \(\mathbf{X}^{(l)} \leftarrow (1-\beta) \cdot \mathbf{X}' + \beta \cdot \mathbf{X}\) \hfill // Residual mixing
    \ENDFOR
    \RETURN \(\mathbf{H} \leftarrow \mathbf{X}^{(h)}\).
\end{algorithmic}
\end{algorithm}

\iffalse
% 5.3 收敛性分析不必要
\begin{theorem}[Convergence of Higher-Order Attribute Propagation]\label{theo:convergence}
Let \(\mathbf{\tilde{A}}\) be a normalized adjacency matrix satisfying \(\|\mathbf{\tilde{A}}\| \le 1\) (e.g., in terms of the spectral norm or row-sum norm), and let the residual coefficient \(\beta \in (0,1]\). Define the iterative update
\[
\mathbf{X}^{(l)} = (1 - \beta)\,\mathbf{\tilde{A}}\,\mathbf{X}^{(l-1)} + \beta\,\mathbf{X},\quad \mathbf{X}^{(0)} = \mathbf{X},
\]
Then the sequence \(\{\mathbf{X}^{(l)}\}\) converges to the unique fixed point
\[
\mathbf{X}^* = \beta\,\left(\mathbf{I} - (1-\beta)\,\mathbf{\tilde{A}}\right)^{-1}\mathbf{X}.
\]
\end{theorem}

\begin{proof}
The detailed proof of the theorem is put in our Supplementary Material due to the space limits.
\end{proof}
 \fi

\begin{theorem}[Heterophily Alleviation via Residual Propagation \cite{geomix}]
\label{theo:heterophily}
Consider a heterophilic graph with \( C \) classes, where each node \( u \) has a class label \( y_u \). Let \( \boldsymbol{\mu}_{c} \) denote the mean feature vector of class \( c \). Suppose the node feature update follows Eq.~\eqref{eq:resX}.  Let
\[
p_0 \;\approx\; \mathbb{P}\bigl(y_v = y_u \;\big|\; v \in \mathcal{N}(u)\bigr)
\]
indicate the heterophily level (with lower \( p \) implying stronger heterophily). Then, after \( l \) propagation steps, the expected feature representation of node \( u \) is
\begin{equation}
\label{eq:expected_alignment}
\mathbb{E}\bigl[\mathbf{x}_{u}^{(l)}\bigr] \;=\; p_{l} \,\boldsymbol{\mu}_{y_u} \;+\;\Bigl(1 - p_{l}\Bigr)
\sum_{c \neq y_u} \frac{1}{C - 1} \,\boldsymbol{\mu}_c,
\end{equation}

where $p_{l} \;=\; \beta \;+\; (1-\beta)\,p_{l-1}.$ As \( l \to \infty \), we have \( p_{l} \to 1 \), Thus, the residual propagation progressively amplifies the influence of the correct class center \( \boldsymbol{\mu}_{y_u} \), thereby mitigating feature contamination from heterophilic neighbors.

\end{theorem}

\begin{proof}
    Since the proof of this theorem directly comes from \cite{geomix}, we omit it for brevity. 
\end{proof}
\vspace{-0.2cm}
\stitle{Balancing Propagation and Feature Retention.}  
Although multi-hop propagation can expand a node’s receptive field and capture global signals, the parameter \(\beta\) in Eq.~\eqref{eq:resX} limits the amount of new information integrated at each step. As indicated in Theorem~\ref{theo:heterophily}, this residual mixing gradually strengthens the alignment with the correct class center, thereby mitigating feature contamination from dissimilar neighbors. This prevents oversmoothing, where nodes of different classes become too similar, but it also sets a limit on how far multi-hop expansion can reach when certain subgraphs are disconnected. In Fig.~\ref{fig:intro}, nodes \(v_1\) and \(v_3\) are directly connected to \(v_2\), so multi-hop propagation allows them to share partial attributes. However, other nodes such as \(v_7\) that are several hops away from \(v_2\) is difficult to effectively enriched by TEA. Consequently, TEA helps some nodes discover shared features (e.g., \(\mathbf{x}_{v_1}\) and \(\mathbf{x}_{v_3}\)), but more distant nodes with no bridging path may also remain overlooked.

\vspace{-0.3cm}
\subsection{Attribute-informed Topology Augmentation}
\label{sec:attribute-informed}
\begin{figure}[t]
 \centering
\includegraphics[width=0.45\textwidth]{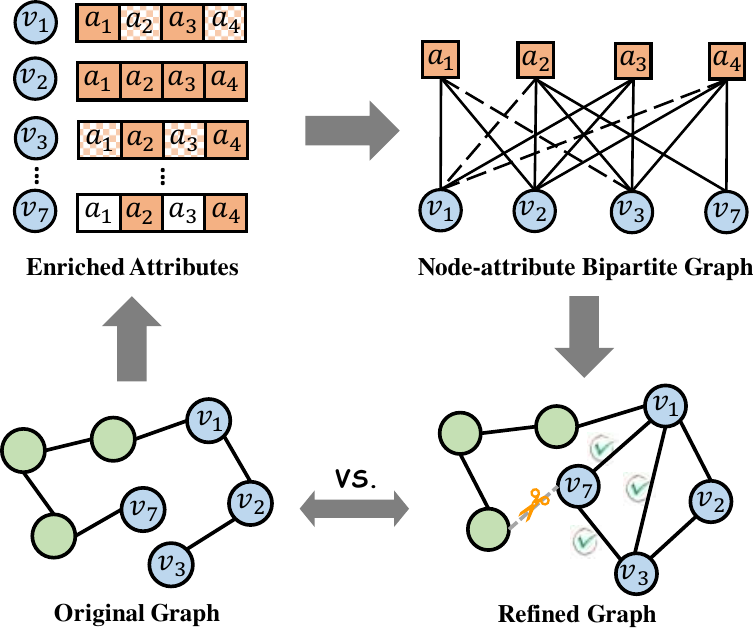}
\caption{Nodes \(v_1, v_2, v_3, v_7\) share the same class. AIT connects distant nodes (e.g., \(v_7\)) via multi-hop feature walks.}
\vspace{-0.5cm}
\label{fig:ait2}
\end{figure}

After obtaining topology-enriched attributes (Section~\ref{sec:topology-enriched}), we further leverage them to refine the original graph structure. Although node features often contain valuable clues about node similarity, traditional distance metrics (e.g., Cosine or Euclidean) can fail under noise or sparsity, especially for pairs of nodes that are multi-hop away. For example, in Fig.~\ref{fig:ait2}, nodes \(v_1\) and \(v_3\) might be recognized as similar after TEA, yet \(v_7\) being too distant from \(v_2\), still remains disconnected. A simple cosine measure on \(\{\mathbf{x}_{v_1},\mathbf{x}_{v_7}\}\) could report near-zero overlap if their local features have not been sufficiently propagated. To address this shortfall, we propose constructing a \emph{node-attribute bipartite graph} and applying the well-known proximity metric PPR to it, thus capturing indirect multi-hop attribute relationships from a \emph{global perspective}. As a result, nodes can connect via shared feature dimensions even if they reside in different parts of the original graph. 

\stitle{Node-attribute Bipartite Graph Construction.} Let \(\mathbf{H} \in \mathbb{R}^{n \times k}\) be the enriched node-feature matrix by TEA (Algorithm \ref{alg:augment_fixed}) in Section \ref{sec:topology-enriched}. We construct the node-attribute bipartite graph as follows (Fig.~\ref{fig:ait2}). Let \( G_b = (U \cup V, E_b) \) be a bipartite graph, in which \( U = \{a_1, a_2, \dots, a_{k}\} \) is the attribute set  and \( V = \{v_1, v_2, \dots, v_{n}\} \) is the node set of the original input graph $G$. The edge set $E_b\subseteq U \times V$ and each edge $(a_i, v_j) \in E_b$ is associated with a non-negative weight $w(a_i, v_j) = H_{ij}$. 
\iffalse
\begin{equation}\label{eq:bipartite-edges}
  E = \left\{ (u_i, v_j) \;\middle|\; H_{i,j} \neq 0 \right\}, \quad w(u_i, v_j) = |H_{i,j}|,
\end{equation}
where \( w(\cdot, \cdot) \) denotes the non-negative weight function that reflects the magnitude of feature \( j \) for node \( i \). \fi
To ensure the robustness of attribute information and prevent spurious signals from negatively impacting similarity computations in practice, we preprocess the feature matrix \( \mathbf{H} \) using the following procedure: \emph{(i)} {Zero-Dimension Removal:} Remove any feature dimension (i.e., column) that is entirely zero across all nodes. \emph{(ii)} {Singleton Filtering:} Optionally filter or down-weight feature dimensions that are nonzero for only a single node, since these features do not contribute to similarity between nodes. By projecting the enriched feature matrix \( \mathbf{H} \) into this bipartite graph, random walks can move from an attribute  \( a_i \) to a node \( v_j \) if $H_{ij}\neq 0$, thereby capturing latent attribute overlaps. The bipartite construction can effectively model such relationships, even in the presence of noise or missing features.

\stitle{Push-based PPR on Bipartite Graphs.}
Since conventional similarity metrics, such as cosine or Euclidean distance, often struggle to capture indirect relationships in the presence of sparse or noisy node features \cite{simpgnn}, we instead adopt the PPR metric, which effectively captures multi-hop connectivity within the feature space.
Recall that  PPR on a graph $G = (V, E)$ with adjacency matrix \( \mathbf{A} \) is defined as the unique stationary distribution \( \boldsymbol{\pi} \) satisfying:
\begin{equation}\label{eq:ppr-standard}
  \boldsymbol{\pi} = \alpha \mathbf{e}_s + (1-\alpha) \boldsymbol{\pi} \mathbf{D}^{-1} \mathbf{A},
\end{equation}
where \( \mathbf{e}_s \) is a one-hot vector indicating the source node \( s \), \( \alpha \in (0,1) \) is the teleportation probability, and \( \mathbf{D} \) is the diagonal degree matrix. Note that Eq.~\eqref{eq:ppr-standard} is a variant of Eq.~\eqref{formula1} \cite{DBLP:conf/focs/AndersenCL06}.

However,  PPR is designed for general unipartite graphs and does not inherently consider the distinctive properties of bipartite graphs. When applied to the bipartite graph \( G_b \), it generates distributions over both \( U \) and \( V \). To address this limitation, we propose a novel and efficient push-based strategy (Algorithm \ref{alg:push}) that retains the final PPR scores exclusively within \( V \), which is our significant technical contribution. Specifically, the procedure initializes the approximate PPR and residue vectors as $\boldsymbol{\hat{\pi}_s} = 0$ and $\boldsymbol{r_s} = \boldsymbol{e_s}$ (Line 1). It then performs forward-push operations on nodes \( v \) that satisfy the condition \( \frac{\boldsymbol{r_s}(v)}{d(v)} > r_{\max} \) (Lines 2-3). Here, \( r_{\max} \) serves as a parameter that balances the trade-off between the precision and computational efficiency of the PPR computation. During each push operation, a fraction \(\alpha\) of the residue is retained, while the remaining \(1 - \alpha\) is propagated to neighboring nodes (Lines 5-8). Once all neighbors of \( v \) have been updated, the residue \(\boldsymbol{r_s}(v)\) is reset to zero (Line 4). Subsequently, the residues accumulated at nodes in \( N(v) \) are pushed back to \( V \), ensuring no residue loss (Lines 9-14). The algorithm terminates once all nodes \( v \in V \) meet \( \frac{\boldsymbol{r_s}(v)}{d(v)} \le r_{\max} \).

\iffalse
Specifically, let \(\boldsymbol{r}\) be a residue vector and \(\boldsymbol{\hat{\pi}}\) the approximate PPR score vector, both initialized to zero, except for \(\boldsymbol{r}(s) = 1\) at the source node \(s\). We iteratively push the residue from nodes in \(V\) to their neighbors in \(U\), and then from those neighbors back into \(V\). Concretely, if at any node \(v \in V\), the ratio \(\boldsymbol{r}(v)/d(v)\) exceeds a threshold \(r_{\max}\), we apply:  
\begin{align}
  \boldsymbol{\hat{\pi}}(v) &\leftarrow \boldsymbol{\hat{\pi}}(v) + \alpha \cdot \boldsymbol{r}(v), \label{eq:ppr-push-1} \\
  \boldsymbol{r}(a) &\leftarrow \boldsymbol{r}(a) + (1-\alpha) \cdot \frac{w(u, v) \cdot \boldsymbol{r}(u)}{\sum_{v_i \in N(u)}{w(u, v_i)}} \quad \forall\, v \in N(u). \label{eq:ppr-push-2}
\end{align}
then reset \(\boldsymbol{r}(u)\) to zero. By adjusting \(r_{\max}\), we can balance runtime efficiency—where a larger threshold reduces the number of push operations—against approximation accuracy, as a smaller threshold yields more precise scores.  
Throughout these updates, only a subset of the relative rankings of PPR values is required, allowing us to discard small residues early for practical performance benefits.\fi

\stitle{Remark.} %This reconstruction process (i.e., Algorithm \ref{alg:push}) benefits significantly from the enriched features obtained in Section~\ref{sec:topology-enriched}. Specifically, by using multi-hop propagation and residual mixing, the topology-enriched attributes already help address local inconsistencies or partial noise. Thus, the push-based PPR  (i.e., Algorithm \ref{alg:push}) over the bipartite graph can leverage these refined features to uncover \emph{global} topological connections, traversing multiple feature dimensions or distant nodes. Consequently, even if some nodes are only partially corrected in the previous step (Section~\ref{sec:topology-enriched}), they can still reach more suitable neighbors via cross-feature walks.
%Besides, our PPR approach captures multi-hop attribute relationships, overcoming sparsity limitations of direct similarity measures. Moreover, with higher-order relationships beyond local structural limits, bipartite PPR will help capture \textit{global} topological information.%在这里区分high-order是指发现了潜在的节点特征之间的关系，global则更强调拓扑上的全局性。
This reconstruction process (Algorithm~\ref{alg:push}) benefits from the enriched features in Section~\ref{sec:topology-enriched}, since multi-hop residual propagation already reduces local inconsistencies. Hence, the push-based PPR on the bipartite graph can further uncover \emph{global} relationships through cross-feature walks. Even if certain nodes are only partially corrected by TEA, bipartite PPR can connect them to more suitable neighbors. Moreover, Theorem~\ref{thm:bppr} provides a lower bound on bipartite PPR scores for nodes reachable via length-\(2c\) paths, ensuring these multi-hop connections are not overlooked. This approach overcomes the sparsity of direct similarity measures, revealing high-order attribute links beyond local structural limits.

\begin{theorem}[Bipartite-PPR lower bound]\label{thm:bppr}
For any node $v_j$ reachable from $s$ by a length-$2c$ path in $G_b$,
\[
\pi_s(v_j)
\;\ge\;
\alpha(1-\alpha)^{2c}\,
\bigl|\mathcal{P}_{2c}(s\!\leadsto\!v_j)\bigr|\,
\Bigl(\tfrac{w_{\min}^2}{d_{\max}^V d_{\max}^U}\Bigr)^{c}.
\]
\end{theorem}

\begin{proof}
The detailed proof of the theorem is put in our Supplementary Material due to the space limits.
\end{proof}

\begin{algorithm}[t] \footnotesize
\caption{\textit{Intra-Set Forward Push over Bipartite Graphs}}
\label{alg:push}
\begin{flushleft}
\textbf{Input}: Bipartite graph $G_b=(U \cup V, E_b)$, source node $s\in V$, teleportation probability $\alpha$, threshold $r_{max}$. \\
\textbf{Output}: $\{ \boldsymbol{\hat{\pi}_s}(v) \mid v \in V \}$.
\end{flushleft}
\begin{algorithmic}[1] %[1] enables line numbers
    \STATE $\boldsymbol{\hat{\pi}_s} \leftarrow \textbf{0}, \boldsymbol{r_s} \leftarrow \boldsymbol{e_s}$;
    \WHILE{$\exists v \in V$ s.t $\frac{\boldsymbol{r_s}(v)}{d(v)} \textgreater r_{max} $}
        \STATE \textit{pick an arbitrary node $v$ with} $\frac{\boldsymbol{r_s}(v)}{d(v)}\textgreater r_{max} $;
        \STATE $r \leftarrow \boldsymbol{r_s}(v),\boldsymbol{r_s}(v) \leftarrow 0 $;
        \STATE $\boldsymbol{\hat{\pi}_s}(v) \leftarrow \boldsymbol{\hat{\pi}_s}(v) + \alpha \cdot r$;
        \FOR{$u \in N(v)$}
            \STATE $\boldsymbol{r_s}(u) \leftarrow \boldsymbol{r_s}(u) + \frac{(1-\alpha) \cdot w(u, v)}{\sum_{u_i \in N(v)}{w(v, u_i)}} \cdot r$;
        \ENDFOR
        \FOR{$u \in N(v)$}
            \STATE $r \leftarrow \boldsymbol{r_s}(u),\boldsymbol{r_s}(u) \leftarrow 0 $;
            \FOR{$v_i \in N(u)$}
                \STATE $\boldsymbol{r_s}(v_i) \leftarrow \boldsymbol{r_s}(v_i) + \frac{w(v_i, u)}{\sum_{\bar{v} \in N(u)}{w(\bar{v}, u)}} \cdot r$;
            \ENDFOR
        \ENDFOR
    \ENDWHILE
    \RETURN $\{ \boldsymbol{\hat{\pi}_s}(v) \mid v \in V \}$;
\end{algorithmic}
\end{algorithm}

\stitle{Co-Augmented Global Graph Reconstruction.} After obtaining node-level PPR scores $\hat{\pi}_{ij}$ among nodes in $V$, a new Co-Augmented global adjacency matrix \( \mathbf{\tilde{A}^{(c)}} \) can be reconstructed using two flexible strategies:\noindent{\textcircled{1} Fixed-size $K$NN Graph:} This method selects a global $K$ such that the resulting graph has a comparable number of edges to the original. For each node $v_i$, the top-$K$ neighbors $v_j$ with the highest $\hat{\pi}_{ij}$ values are selected to form edges in the new adjacency matrix. %By matching $K$ to the target edge count, this approach maintains a similar density while focusing on the most relevant connections. In practice, we set $K$ as the average degree of the original graph. Once PPR scores are sorted for each node, constructing the graph involves retaining only the top-$r$ entries.
\noindent{\textcircled{2} Edge Addition/Removal:}  
Alternatively, the original adjacency matrix can be incrementally modified based on the PPR ranking. For each node, edges are selectively added or removed based on their PPR ranking. A fixed number of edges can be added or removed per node (e.g., $k_{\mathrm{add}}$, $k_{\mathrm{del}}$) to update the original adjacency matrix. %These values are typically set between 1 and 3 to avoid significant changes in the graph's edge count while selectively enhancing connectivity or pruning noisy links.In both strategies, only the \emph{relative} ordering of PPR scores is necessary, allowing a larger threshold $r_{\max}$ in the forward-push procedure to optimize computational efficiency. 

Overall, by combining the enriched attributes (Section~\ref{sec:topology-enriched}) with the bipartite PPR process here, we achieve a robust co-augmentation frame that captures both localized and global signals, effectively mitigating noise or partial inconsistencies in real-world graphs.

% ----------- 4.3 Integration: Dual-Channel CoATA GNN -----------
\subsection{Dual-Channel GNN with  Alignment}
\label{sec:4.4}

%This design is inspired by recent dual-view GNNs~\cite{dualviewgnn} but specifically adapted to our co-augmentation pipeline. 
Having introduced two complementary graph views in Sections~\ref{sec:topology-enriched} and \ref{sec:attribute-informed}, we now unify them using a dual-channel scheme. The original adjacency matrix \(\mathbf{\tilde{A}}\) retaining raw but possibly noisy edges, thereby preserving faithful local patterns. In contrast, the refined adjacency matrix \(\mathbf{\tilde{A}^{(c)}}\)  derived via bipartite PPR, captures broader global connectivity that may be missed by the original graph alone.

\vspace{0.5em}
\noindent
\textbf{Dual-Channel Model Setup.}\;
We employ a single GNN architecture with \emph{shared} parameters across both channels. The original channel processes \(\bigl(\mathbf{\tilde{A}}, \mathbf{X}\bigr)\), while the co-augmented channel takes \(\bigl(\mathbf{\tilde{A}^{(c)}}, \mathbf{X}\bigr)\), producing:
\begin{align}
\hat{\mathbf{Y}} = \mathrm{GNN}\bigl(\mathbf{\tilde{A}}, \mathbf{X}\bigr), \quad
\hat{\mathbf{Y}}' = \mathrm{GNN}\bigl(\mathbf{\tilde{A}^{(c)}}, \mathbf{X}\bigr),
\end{align}

%Both yield an output dimension \(\mathbb{R}^{n\times C}\). 
By sharing weights, we reduce the overall number of parameters and encourage robustness across both original and augmented structures. Each channel is supervised with a cross-entropy term:
\begin{align}
\label{eq:CE-orig}
\mathcal{L}_{ce} \;=\;
-\sum_{\,i \in \mathcal{V}_l}\,
\mathbf{Y}(i)\,\log\,\hat{\mathbf{Y}}(i),\
\mathcal{L}_{ce}'
\;=\;
-\sum_{\,i \in \mathcal{V}_l}\,
\mathbf{Y}(i)\,\log\,\hat{\mathbf{Y}}'(i),
 \end{align}
Where $\mathbf{Y}(i)$ denotes the label for node $i$, and $\mathcal{V}_l$ is the training set.

%Optimizing these two terms jointly prevents reliance on only one adjacency, thereby mitigating incomplete or noisy links. To speed training, we typically reuse the same optimizer step for both channels in each epoch, and adopt an early-stopping criterion based on validation accuracy.

\stitle{Consistency Across Multiple Augmentations.}
In practice, we construct two specific augmented graphs \( \mathbf{\tilde{A}^{(c)} }\): one using the Fixed-size \( K \)NN Graph strategy and the other using the Edge Addition and Removal strategy (see Section~\ref{sec:attribute-informed}). %These two augmentation methods complement each other: \( K \)NN ensures that each node maintains strong global connections, while edge addition/removal refines the original structure by selectively enhancing or pruning links based on PPR scores.
To ensure coherence and leverage both augmented graphs effectively, we adopt a consistency loss that aligns the model’s outputs.
\begin{align}\label{eq:feature-consistency-mod}
\mathcal{L}_{co}
=
\frac{1}{S}\sum_{s=1}^{S}
\sum_{i\in V}
\Bigl\|
\,\mathbf{Y}^{\mathrm{agg}}(i)
\;-\;
\hat{\mathbf{Y}}'_s(i)
\,\Bigr\|_2^2,
\end{align}
where \(\hat{\mathbf{Y}}'_s\) denotes the augmented-channel predictions under the $s$-th augmented adjacency matrix, and \(\mathbf{Y}^{\mathrm{agg}}(i)\) is an aggregated pseudo-label \cite{pl} (e.g., averaged predictions). This term encourages outputs to remain consistent across different $\mathbf{\tilde{A}^{(c)}}$.

\stitle{Dual-Channel Prototype Alignment.} Even if the augmented graphs are consistent internally, they may still diverge from the original adjacency matrix in terms of class-level semantics. Inspired by \cite{progcl,plabels}, we impose a \emph{prototype alignment} to align embeddings \emph{across} channels. Let \(\mathbf{Z}, \mathbf{Z}' \in \mathbb{R}^{n\times d}\) be the hidden embeddings for the original and augmented channels, respectively. For class \(j\in\{1,\dots,C\}\), we define:
\begin{align}
\mathbf{p}_j
=
\frac{\sum_{\,i:\,\arg\max\,\hat{\mathbf{Y}}_i = j}
\,t_i\,\mathbf{Z}_i}
{\sum_{\,i:\,\arg\max\,\hat{\mathbf{Y}}_i = j}t_i},
\quad
\mathbf{p}_j'
=
\frac{\sum_{\,i:\,\arg\max\,\hat{\mathbf{Y}}'_i = j}
\,t_i\,\mathbf{Z}'_i}
{\sum_{\,i:\,\arg\max\,\hat{\mathbf{Y}}'_i = j}t_i},
\end{align}
where \(t_i\) is a confidence weight (1 if labeled, or the highest predicted probability if unlabeled). We then adopt a contrastive loss:
\begin{equation}\small
\mathcal{L}_{dpa}
\,=\,
-\frac{1}{2c}
\sum_{j=1}^c
\left(
\log
\frac{f(\mathbf{p}_j, \mathbf{p}_j')}
{\sum_{q\neq j}f(\mathbf{p}_j,\mathbf{p}_q')}
+
\log
\frac{f(\mathbf{p}_j, \mathbf{p}_j')}
{\sum_{q\neq j}f(\mathbf{p}_q,\mathbf{p}_j')}
\right),
\end{equation}
with \(f(\mathbf{u},\mathbf{v}) = e^{\cos(\mathbf{u},\mathbf{v})/\tau}\), where \(\tau\) is typically set to 0.5. This alignment pulls prototypes of the same class closer and pushes apart those from different classes, improving semantic consistency and reducing label mismatch~\cite{clmit}, as stated in the following theorem.

\begin{theorem}[Prototype Alignment Guarantee]\label{thm:proto_align}
Minimizing the contrastive loss \(\mathcal{L}_{dpa}\) ensures that prototypes from the same class across channels are pulled closer and those from different classes remain separated.
\end{theorem}

\begin{proof}
The detailed proof of the theorem is put in our Supplementary Material due to the space limits.
\end{proof}

\stitle{Overall Objective.} We combine all the above loss components:
\begin{align}
\label{eq:total-loss-new}
\mathcal{L}
\,=\,
\mathcal{L}_{ce} + \lambda_1\,\mathcal{L}_{ce}'
\;+\;
\lambda_2\,\mathcal{L}_{co}
\;+\;
\lambda_3\,\mathcal{L}_{dpa},
\end{align}
where \(\lambda_1,\lambda_2,\lambda_3\) modulate the trade-off among supervision, inter-augmentation consistency, and cross-channel alignment.

%As a thus, our CoATA exploits a single parameterized GNN on two distinct adjacency matrices, regularizing the outputs through multi-graph consistency and class-level prototype alignment. Empirical results confirm that combining original and augmented structures enhances robustness to noise and missing edges, while adding negligible overhead compared to fully separate GNNs.

\section{Empirical Results}
\label{sec:experiments}

\iffalse
In this section, we evaluate the performance of our CoATA model on semi-supervised node classification, focusing on the effectiveness of CoATA. To guide our analysis, we organized the discussion according to three key research questions.

\noindent\textbf{RQ1.} Does CoATA outperform state-of-the-art baselines on node classification tasks?\\
\noindent\textbf{RQ2.} Do all the proposed adaptive graph augmentation schemes topology-enriched attribute (\textit{TEA}) and attribute-informed topology (\textit{AIT}) improve our model’s performance? How does each component specifically contribute?\\
\noindent\textbf{RQ3.} Is CoATA sensitive to hyperparameters? How do key hyperparameter choices affect model accuracy?
\fi

\subsection{Experimental Setup}

\noindent\textbf{Datasets.} We evaluate our solutions on seven standard graph benchmarks (Table \ref{tab:dataset}): three citation networks (Cora, CiteSeer, PubMed~\cite{cora}), two co-authorship networks (Coauthor-CS, Coauthor-Phy~\cite{coauthor}), and two heterophilic graphs (Squirrel, Chameleon~\cite{heterograph}). For citation networks, we adopt the data splits of~\cite{cora}, using 20 labeled nodes per class for training, 500 for validation, and 1000 for testing. For co-authorship networks, we follow~\cite{coauthor} with 20 labeled nodes per class for training, 30 for validation, and the rest for testing. For Squirrel and Chameleon, we follow the recent work ~\cite{heterograph} which filters out the overlapped nodes in the original datasets and uses its provided data splits. A detailed description of datasets is put in our Supplementary Material due to the space limits.

\noindent\textbf{Baselines.}
We compare our proposed solutions with eleven competitors, which are summarized into three groups as follows. (1) Vanilla: GCN \cite{gcn}, GAT \cite{gat}, SGC \cite{sgc}, and APPNP \cite{ppnp}; (2) Topology-oriented: S\textsuperscript{3}\text{-CL} \cite{s3cl}, GloGNN \cite{glognn}, and PSAGNN \cite{psagnn}; (3) Attribute-oriented: Mixup \cite{mixup}, Geomix \cite{geomix}, GraphMix \cite{graphmix}, and Simp-GCN \cite{simpgnn}. To mitigate the impact of randomness, we repeat each method five times and report the average value as the final performance.

\noindent\textbf{Parameters and Implementations.} 
 Unless stated otherwise, we use the corresponding default parameters for all baselines. For our model, the key hyperparameters are $\alpha$, $\beta$, and $ h$, which are set within the ranges $\alpha$ in [0.1, 0.9], $\beta$ in [0.1, 0.9], and $ h $ in [1, 4]. Our implementation uses Python, with DGL and PyG as the primary libraries. Our Supplementary Material provides further implementation details. All experiments are conducted on an Ubuntu 22.04.3 machine equipped with an NVIDIA RTX 4000 GPU, an Intel Xeon(R) 2.20GHz CPU, and 320GB of RAM.

\renewcommand{\arraystretch}{0.9} % 行间距
\begin{table}[t]
\centering
\caption{Statistics of datasets.}
\vspace{-2mm}
\resizebox{1\columnwidth}{!}{
\begin{tabular}{l|cccc}
\toprule
\textbf{Dataset} & \textbf{\#Nodes} & \textbf{\#Edges} & \textbf{\#Classes} & \textbf{\#Features}\\
\midrule
Cora              & 2,708    & 5,429     & 7       & 1,433  \\ 
CiteSeer          & 3,327    & 4,732     & 6       & 3,703  \\ 
PubMed            & 19,717   & 44,338    & 3       & 500    \\ 
\midrule
Coauthor-CS          & 18,333   & 163,788   & 15      & 6,805  \\ 
Coauthor-Phy  & 34,493   & 495,924   & 5       & 8,415  \\
\midrule
Chamelon          & 2,277    & 36,101    & 5       & 2,325  \\ 
squirrel          & 5,201    & 217,073   & 5       & 2,089  \\ 
\bottomrule
\end{tabular}
}
\label{tab:dataset}
\vspace{-2mm}
\end{table}

\subsection{Effectiveness Evaluation}

\begin{table*}[htbp]
\centering
\small                   % 调整字体大小，可以使用 \footnotesize, \small, \normalsize 等
\setlength{\tabcolsep}{5pt}  % 控制列与列之间的距离（默认一般为6pt左右）
\renewcommand{\arraystretch}{1.1}  % 控制表格的行间距，数值越大行距越大
\caption{Node classification accuracy (\%). Each result is presented as the average $\pm$ standard deviation, with the best result highlighted in \textbf{bold}.}\vspace{-0.3cm}
\label{trans}
\begin{tabular*}{0.95\textwidth}{lcccccccc}
\toprule
\textbf{Type} & \textbf{Method} & \textbf{Cora} & \textbf{PubMed} & \textbf{Citeseer} & \textbf{Coauthor-CS} & \textbf{Coauthor-Phy} & \textbf{Squirrel}& \textbf{Chameleon} \\
\midrule
\multirow{4}{*}{Vanilla} 
& GCN        & 81.63 $\pm$ 0.45 & 78.88 $\pm$ 0.65 & 70.98 $\pm$ 0.37 & 91.16 $\pm$ 0.52 & 92.85 $\pm$ 1.03 & 39.47 $\pm$ 1.47 & 40.89 $\pm$ 4.12 \\
& GAT        & 82.98 $\pm$ 0.88 & 78.58 $\pm$ 0.52 & 72.20 $\pm$ 0.99 & 90.57 $\pm$ 0.37 & 92.70 $\pm$ 0.58 & 35.96 $\pm$ 1.73 & 39.29 $\pm$ 2.84 \\
& SGC        & 80.35 $\pm$ 0.24 & 78.75 $\pm$ 0.17 & 71.87 $\pm$ 0.14 & 90.37 $\pm$ 1.01 & 92.80 $\pm$ 0.15 & 39.04 $\pm$ 1.92 & 39.35 $\pm$ 2.82 \\
& APPNP      & 83.33 $\pm$ 0.52 & 79.78 $\pm$ 0.66 & 71.83 $\pm$ 0.52 & 91.97 $\pm$ 0.33 & 93.86 $\pm$ 0.33 & 37.64 $\pm$ 1.63 & 38.25 $\pm$ 2.83 \\
\midrule
\multirow{3}{*}{Topology} 
&{S\textsuperscript{3}-CL}    & 84.83 $\pm$ 0.21 & 80.73 $\pm$ 0.28 & 74.97 $\pm$ 0.52 & 91.29 $\pm$ 0.17 & 93.26 $\pm$ 0.42 & 38.13 $\pm$ 1.36 & 40.33 $\pm$ 3.21 \\
& GloGNN     & 82.31 $\pm$ 0.42 & 79.83 $\pm$ 0.21 & 72.16 $\pm$ 0.64 & 90.82 $\pm$ 0.45 & 92.79 $\pm$ 0.67 & 35.77 $\pm$ 1.32 & 26.17 $\pm$ 4.33 \\
& PSAGNN     & 82.91 $\pm$ 0.79 & 79.39 $\pm$ 1.25 & 71.18 $\pm$ 0.32 & 90.52 $\pm$ 0.32 & 92.67 $\pm$ 0.59 & 37.82 $\pm$ 1.12 & 40.23 $\pm$ 3.42 \\
\midrule
\multirow{4}{*}{Attribute} 
& Mixup      & 81.84 $\pm$ 0.94 & 79.16 $\pm$ 0.49 & 72.20 $\pm$ 0.95 & 91.36 $\pm$ 0.37 & 93.89 $\pm$ 0.49 & 37.95 $\pm$ 1.52 & 39.56 $\pm$ 3.13 \\
& GeoMix   & 84.22 $\pm$ 0.85 & 80.18 $\pm$ 0.99 & 75.12 $\pm$ 0.26 & 92.23 $\pm$ 0.14 & 94.56 $\pm$ 0.06 & 40.95 $\pm$ 1.12 & 42.67 $\pm$ 2.44 \\
& GraphMix   & 83.80 $\pm$ 0.62 & 79.38 $\pm$ 0.39 & 74.28 $\pm$ 0.45 & 91.89 $\pm$ 0.36 & 94.32 $\pm$ 0.28 & 38.41 $\pm$ 1.36 & 41.75 $\pm$ 3.51 \\
& SimP-GCN   & 82.83 $\pm$ 0.26 & \textbf{80.84 $\pm$ 0.17} & 72.63 $\pm$ 0.59 & 90.54 $\pm$ 0.45 & 93.43 $\pm$ 0.54 & 39.26 $\pm$ 1.87 & 40.93 $\pm$ 4.25 \\
\midrule
\textbf{This paper} & \textbf{CoATA} 
                   & \textbf{85.07 $\pm$ 0.31} 
                   & 80.21 $\pm$ 0.16 
                   & \textbf{75.37 $\pm$ 0.21} 
                   & \textbf{92.31 $\pm$ 0.29} 
                   & \textbf{95.12 $\pm$ 0.27} 
                   & \textbf{42.26 $\pm$ 1.31} 
                   & \textbf{44.07 $\pm$ 4.40} \\
\bottomrule

\end{tabular*}
\end{table*}

Table \ref{trans} compares CoATA with 11 baselines on seven datasets. On homophilous datasets, we have the following observations: (1) CoATA achieves the best performance on four of five homophilous datasets. For example, on \emph{Coauthor-Phy}, our CoATA attains 95.12\%, surpassing all competitors by at least 0.56\%. (2) On \emph{PubMed}, CoATA ranks third and slightly behind the champion SimP-GCN. One possible reason is the relatively low-dimensional features in \emph{PubMed} curtail the TEA’s ability to enhance noisy attributes. On the two heterophilic datasets \emph{Squirrel} and \emph{Chameleon}, we have the following observations. (1) Most topology-centric baselines fail to surpass even a plain GCN, indicating that naive structural modifications can inadvertently reinforce mismatched neighbors. (2) CoATA outperforms all competitors by at least 1.31\% and 1.40\%, respectively, confirming that its dual-channel architecture effectively addresses \emph{heterophily}. %Overall, these findings provide strong evidence that our proposed CoATA’s co-augmentation strategy fosters superior robustness and accuracy in both homophilous and heterophilic graphs when contrasted to baselines.
Overall, CoATA is consistently top on both graph types.

 %Nonetheless, CoATA remains highly competitive overall, underscoring its robust co-augmentation pipeline that blends higher-order feature propagation (\textit{TEA}) with global graph refinement (\textit{AIT}). By simultaneously correcting node attributes and leveraging multi-hop bipartite PPR, CoATA mitigates the risk of propagating erroneous edges or spurious features, particularly in graphs with incomplete or noisy connections. Concretely, \emph{TEA} uses multi-hop residual propagation to smooth only partially among nearby nodes, avoiding over-saturation in areas with strong heterophily; meanwhile, \emph{AIT} constructs a bipartite representation that captures subtle attribute overlaps at a global scale via random walks. Finally, the \textit{DPA} module aligns representations at a class-prototype level, ensuring consistency between raw and augmented views. This synergy is especially beneficial where local edges are unreliable, as it leverages high-order attribute signals without losing crucial node-specific distinctions. 

\begin{table}[t]
    \centering
    \caption{Ablation Studies. $O\textsubscript{1}$ refers to AIT, $O\textsubscript{2}$ represents TEA, and $O\textsubscript{3}$ denotes DPA. \emph{w/o} means without.}
    \vspace{-3mm}
    \resizebox{0.45\textwidth}{!}{
    \begin{tabular}{lcccc}
        \toprule
        \textbf{Datasets}  & \textbf{Citeseer} & \textbf{Coauthor-CS} & \textbf{Chameleon} \\
        \midrule
        $GCN$                   
        & 70.98 $\pm$ 0.37  & 91.16 $\pm$ 0.52  & 40.89 $\pm$ 4.12 \\
        $O_{\textsubscript{1}}$                    
        & 71.07 $\pm$ 0.45  & 91.84 $\pm$ 0.45  & 42.11 $\pm$ 4.52 \\
        $O_{\textsubscript{1}} + O_{\textsubscript{2}}$          
        & 74.90 $\pm$ 0.16  & 92.01 $\pm$ 0.35  & 42.63 $\pm$ 4.42 \\
        $O_{\textsubscript{1}} + O_{\textsubscript{2}} + O_{\textsubscript{3}}$  
        & 75.37 $\pm$ 0.21  & 92.31 $\pm$ 0.29  & 44.07 $\pm$ 4.40 \\
        
        \midrule
        $w/o\ \mathcal{L}_{co}  $   & 72.60 $\pm$ 0.43  & 91.22 $\pm$ 0.43  & 42.66 $\pm$ 4.18 \\
        $w/o\ \mathcal{L}_{ce}' $   & 72.80 $\pm$ 0.24  & 90.53 $\pm$ 0.24  & 40.37 $\pm$ 4.42 \\
        
        \bottomrule
        
    \end{tabular}
    }
    \label{tab:ablation}
\end{table}

\subsection{Ablation Studies}
\label{sub:ablation}

Table~\ref{tab:ablation} shows the performance of each CoATA module on two homophilous datasets \emph{Citeseer} and \emph{CoAuthor-CS}, and one heterophilic dataset \emph{Chameleon}. Other datasets behave similarly and are omitted for brevity. The theoretical rationale behind $O_{\textsubscript{1}}$ (\emph{AIT}) is to refine the graph structure via adjacency augmentation; however, its modest gain (e.g., only +0. 09\% on Citeseer) indicates that structural enhancement alone is insufficient, especially in sparsely connected graphs that highlight the need for attribute enrichment. Incorporating $O_{\textsubscript{2}}$ (\textit{TEA}) facilitates multi-hop neighbor signal propagation and mitigates local feature noise, a claim corroborated by the marked accuracy improvement observed on Citeseer (+ 3.92\%), which aligns with our theoretical expectation that deeper neighborhood interactions strengthen feature propagation. The further addition of $O_{\textsubscript{3}}$ (\textit{DPA}) introduces prototype alignment to reconcile raw and augmented representations, yielding the largest performance boost (+4.39\%), and confirming that aligning multi-view representations is a key to overcoming systematic biases. Moreover, the significant drops observed when omitting consistency regularization (w/o $\mathcal{L}_{co}$) or the augmented-channel supervision (w/o $\mathcal{L}_{ce}'$) further emphasize the theoretical premise that integrated multi-view learning is crucial for improving performance.

\subsection{Visualization of our model \emph{CoATA}}
\begin{figure}[t]
\centering
\subfigure[\emph{GCN}]{\includegraphics[width=0.11\textwidth]{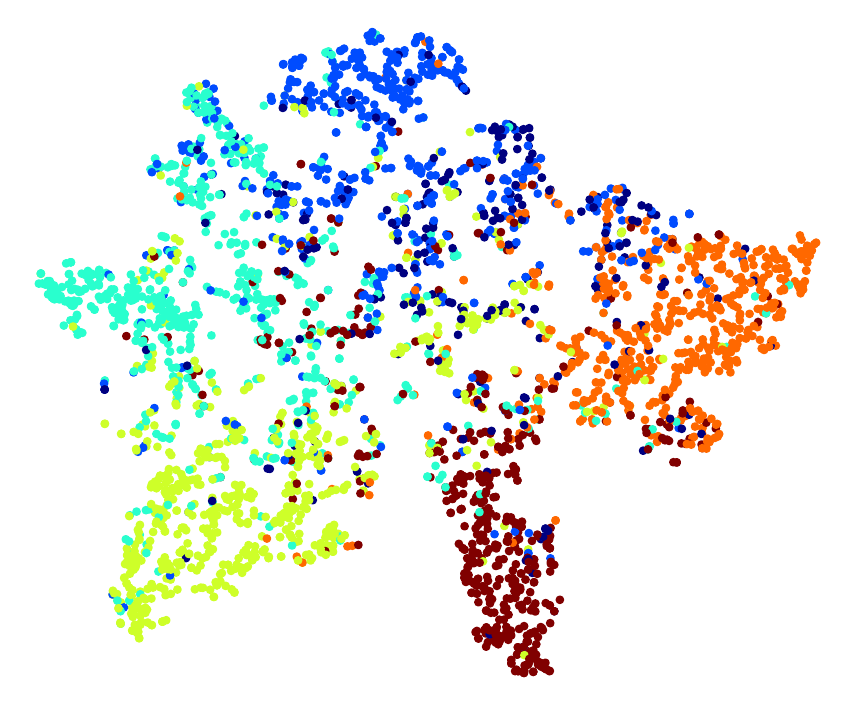}}
\subfigure[\emph{AIT}]{\includegraphics[width=0.11\textwidth]{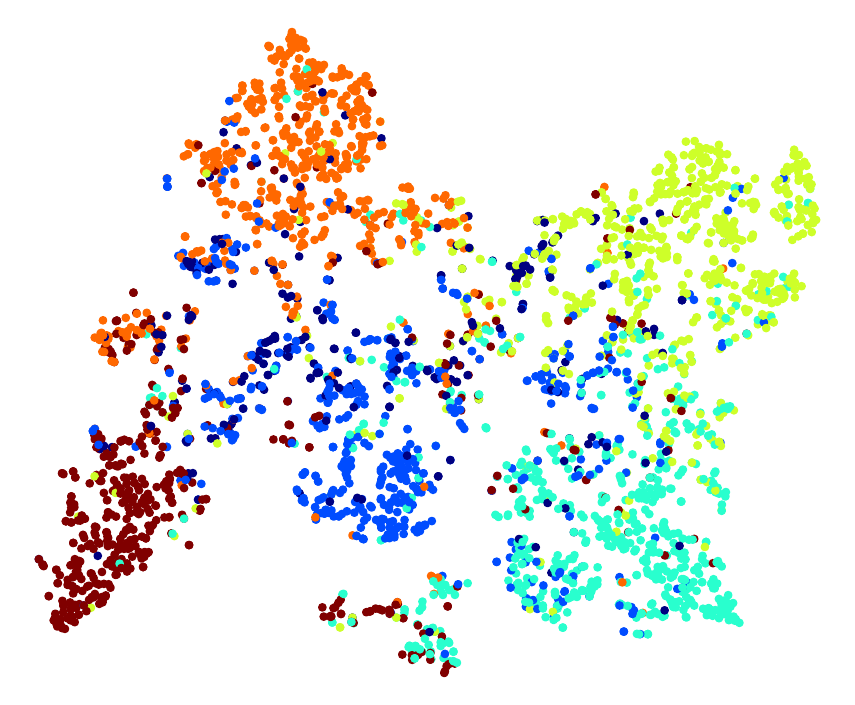}}
\subfigure[\emph{TEA + AIT}]{\includegraphics[width=0.11\textwidth]{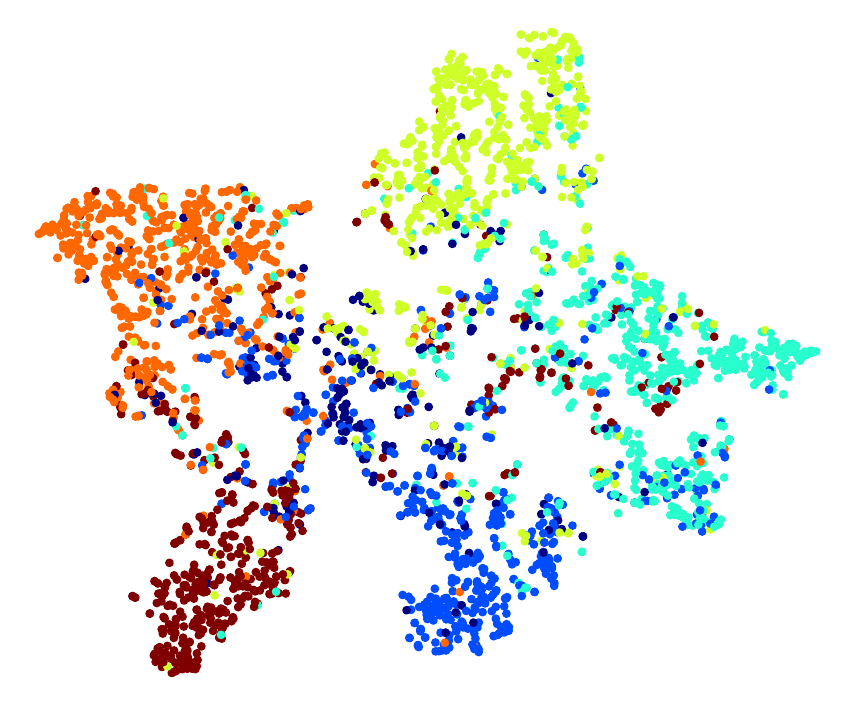}}
\subfigure[\emph{CoATA}]{\includegraphics[width=0.11\textwidth]{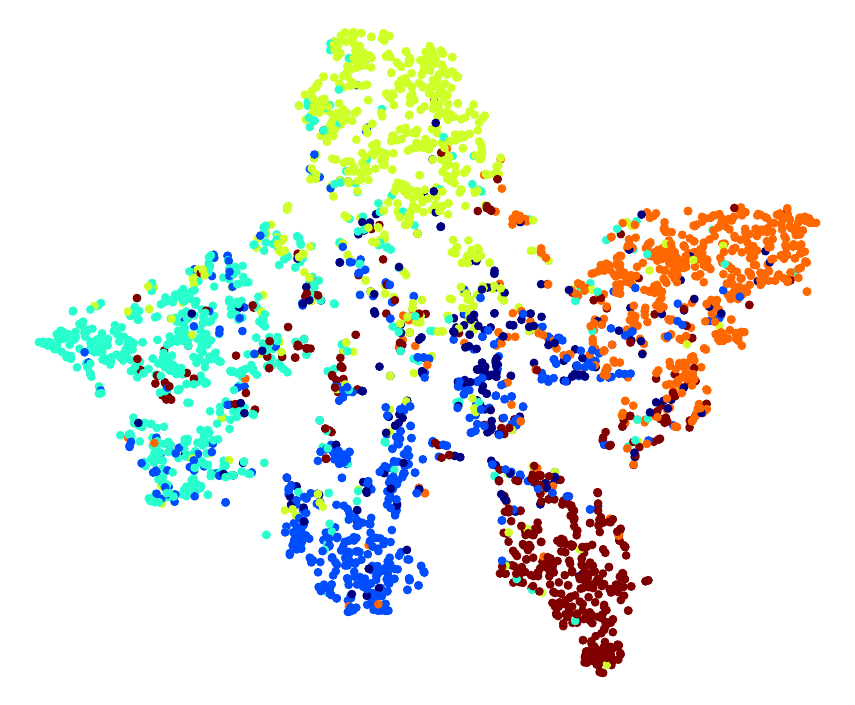}}
\caption{Visualization of our model \emph{CoATA}. Nodes are colored by their corresponding labels.}
\label{fig:visual}
\vspace{-0.5cm}
\end{figure}

Fig.~\ref{fig:visual} visualizes the embedding vectors learned by different methods on the \emph{Citeseer} dataset, with similar trends observed across other datasets, which we omit for brevity. As can be seen, in Fig.~\ref{fig:visual}(a), Vanilla GCN struggles to separate classes, as it aggregates disparate labels due to the limitations of one-hop propagation. In Fig.~\ref{fig:visual} (b), AIT tightens clusters, yet some label overlap persists, highlighting the need for more nuanced feature propagation. In Fig.~\ref{fig:visual}(c), with the introduction of TEA combined with AIT, multi-hop feature propagation further sharpens the decision boundaries, reducing label mixing. Finally, the full CoATA model in Fig.~\ref{fig:visual}(d) achieves the sharpest and most distinct class separation, demonstrating how DPA resolves residual ambiguities by aligning node embeddings at the class level, leading to clearer decision boundaries.

% ------------------------------RQ3 ------------------------------
\subsection{Hyperparameter Analysis}
\label{subsec:RQ3}

\begin{figure}[t]

  \centering
  % 第一行图片
  \begin{minipage}[b]{1\columnwidth}
    \centering
    \includegraphics[width=\linewidth]{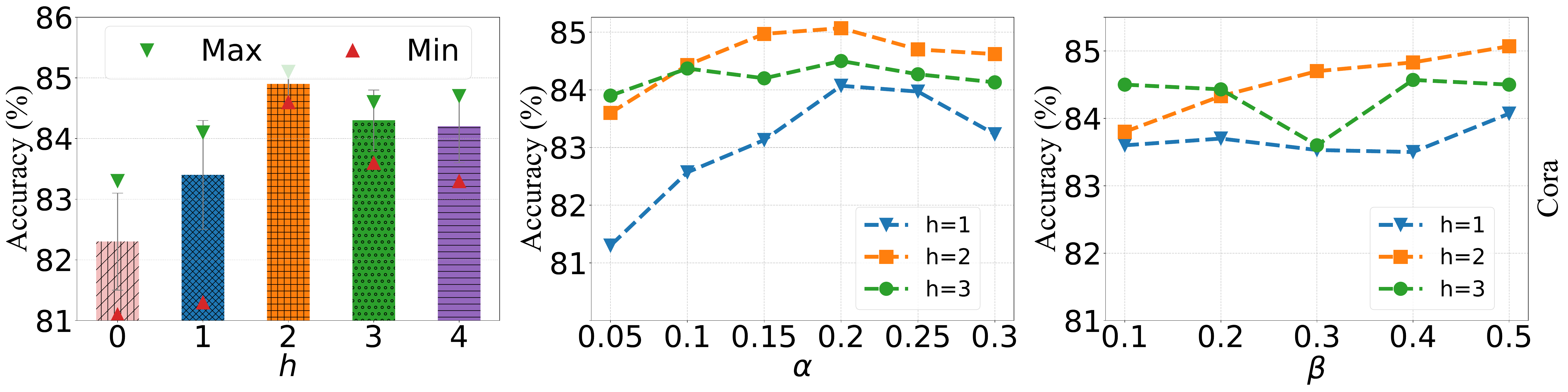}
    \caption*{}
  \end{minipage}  
  \vspace{-14.5mm} % 两行之间的垂直间距
  
  % 第二行图片
  \begin{minipage}[b]{1\columnwidth}
    \centering
    \includegraphics[width=\linewidth]{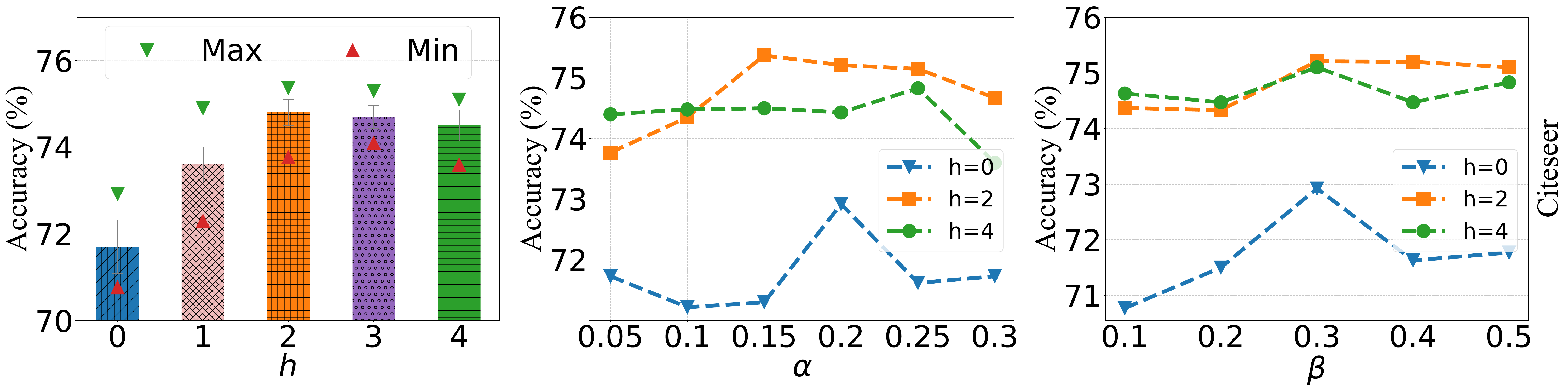}
    \caption*{}
  \end{minipage}
   \vspace{-14.5mm}
   
  % 第三行图片
  \begin{minipage}[b]{1\columnwidth}
    \centering
    \includegraphics[width=\linewidth]{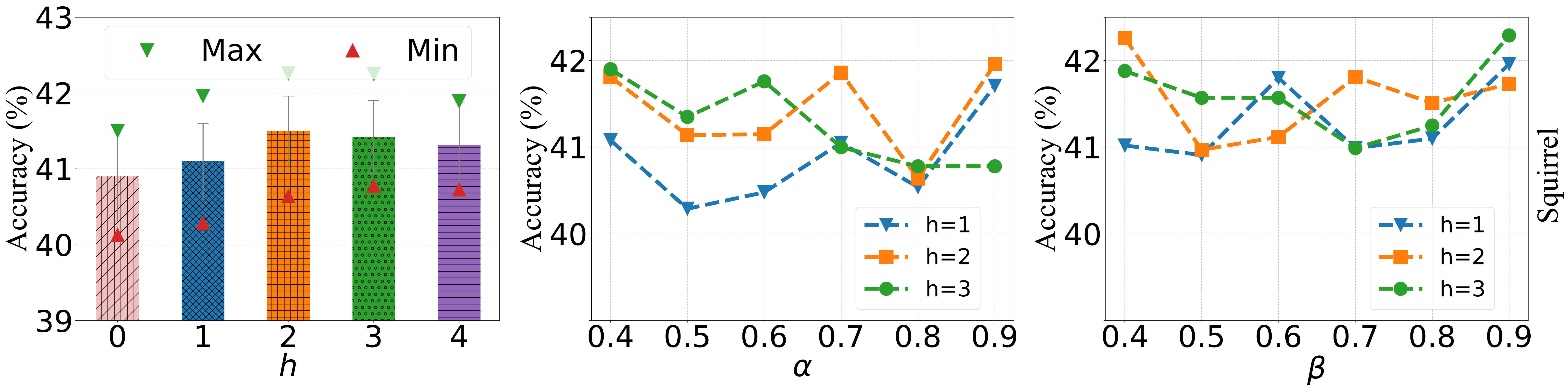}
    \caption*{}
    \vspace{-12.5mm}
    \caption{\small
    \textbf{Hyperparameter Analysis.}
Left bar charts depict average accuracy and standard deviations for $h$, highlighting best/worst results by green/red arrows. Right plots show how varying $\alpha$ and $\beta$ affect performance under different $h$.}
  \label{fig:hy1}
  \end{minipage} 
\end{figure}
Fig.~\ref{fig:hy1} illustrates the impact of three core hyperparameters in CoATA, focusing on \(\alpha\) (teleportation probability in \emph{AIT}), \(\beta\) (residual mixing coefficient in \emph{TEA}), and \(h\) ( propagation depth from Algorithm~\ref{alg:augment_fixed}). 
\stitle{Teleportation Probability \(\alpha\).}
In our push-based PPR formulation Eq.~\eqref{eq:ppr-standard}, \(\alpha\) controls the balance between restarting at the source and propagating information along the bipartite graph edges. On homophilous datasets such as \emph{Cora} and \emph{Citeseer}, moderate values (\(\alpha\approx0.2\)) effectively merge local feature overlaps with global connectivity. Conversely, for heterophilic graphs like \emph{Squirrel}, where local neighborhoods are less reliable, a larger value (\(\alpha\approx0.5\)) is preferable to ensure that even distant yet semantically related nodes contribute to the final similarity scores.

\stitle{Residual Mixing Coefficient \(\beta\).}  
Within the TEA module, Eq.~\eqref{eq:resX} controls the trade-off between multi-hop aggregated features and the original node attributes. As shown in Fig.~\ref{fig:hy1}, in homophilous graphs such as \emph{Cora} and \emph{Citeseer}, the accuracy increases steadily as \(\beta\) rises from 0.1 to around 0.3, indicating that a moderate \(\beta\) effectively fuses local information with higher-order structural signals. In contrast, at \(\beta=0.5\), performance tends to plateau or even decline, suggesting that excessive residual mixing may hinder the beneficial integration of neighborhood information. Meanwhile, in the heterophilic dataset \emph{Squirrel}, the model exhibits robust performance in a wider range of \(\beta\) values, although \(\beta\approx0.5\) appears to be optimal. A broader analysis of \(\beta\) is provided in the Supplementary Material.

\stitle{Propagation Depth \(h\).}
The propagation depth \(h\) determines the number of iterations over which node attributes diffuse across the graph. The bar charts (left column of Fig.~\ref{fig:hy1}) show that on \emph{Cora} and \emph{Citeseer}, setting \(h=2\) or \(3\) improves accuracy by approximately 2\%–3\% compared to \(h=0\), while increasing \(h\) beyond 3 yields diminishing returns or slight drops due to oversmoothing. A similar trend is observed on \emph{Squirrel}, though with higher variance owing to its strongly heterophilic regions. Overall, a moderate depth of \(h=2\) or \(3\) is a robust default, striking a desirable balance between capturing global context and avoiding excessive smoothing.

% In summary, these results demonstrate that a robust default configuration—such as \(\alpha\approx0.2\) for homophilous graphs (and \(\alpha\approx0.5\) for heterophilic graphs), \(\beta\approx0.5\), and a propagation depth \(h\) of 2 or 3 which consistently yields strong performance across diverse datasets. This indicates that CoATA effectively balances local and global signals without requiring extensive hyperparameter tuning.

\section{Conclusion}
\label{sec:conclusion}
In this paper, we introduce CoATA, a novel dual-channel GNN framework designed for Co-Augmentation of Topology and Attribute. This framework refines both topology structures and node attributes, enhancing the robustness of GNNs. Specifically, CoATA first employs the Topology-Enriched Attribute \emph{(TEA)} module to incorporate higher-order structural information into node features, effectively mitigating issues related to noise and sparsity. Subsequently, the Attribute-Informed Topology \emph{(AIT)} module constructs a node-attribute bipartite graph and utilizes the Personalized PageRank proximity metric to reveal multi-hop feature relationships. Following these steps, a Dual-Channel GNN aligns prototypes between the raw and refined graphs. To validate the effectiveness of our proposed solutions, we conducted comprehensive experiments on seven real-world datasets, including five homophilic and two heterophilic graphs, comparing our approach against eleven baselines.

\section{Acknowledgments}
The work was supported by the National Natural Science Foundation of China 62402399. Longlong Lin and Tao Jia are the corresponding authors of this paper.

\clearpage

\bibliographystyle{ACM-Reference-Format}
\balance
\bibliography{myreference}


\begin{thebibliography}{62}

%%% the Web address.

\ifx \showCODEN    \undefined \def \showCODEN     #1{\unskip}     \fi
\ifx \showISBNx    \undefined \def \showISBNx     #1{\unskip}     \fi
\ifx \showISBNxiii \undefined \def \showISBNxiii  #1{\unskip}     \fi
\ifx \showISSN     \undefined \def \showISSN      #1{\unskip}     \fi
\ifx \showLCCN     \undefined \def \showLCCN      #1{\unskip}     \fi
\ifx \shownote     \undefined \def \shownote      #1{#1}          \fi
\ifx \showarticletitle \undefined \def \showarticletitle #1{#1}   \fi
\ifx \showURL      \undefined \def \showURL       {\relax}        \fi
% invisible to TeX
\providecommand\bibfield[2]{#2}
\providecommand\bibinfo[2]{#2}
\providecommand\natexlab[1]{#1}
\providecommand\showeprint[2][]{arXiv:#2}

\bibitem[Andersen et~al\mbox{.}(2006)]%
        {DBLP:conf/focs/AndersenCL06}
\bibfield{author}{\bibinfo{person}{Reid Andersen}, \bibinfo{person}{Fan R.~K. Chung}, {and} \bibinfo{person}{Kevin~J. Lang}.} \bibinfo{year}{2006}\natexlab{}.
\newblock \showarticletitle{Local Graph Partitioning using PageRank Vectors}. In \bibinfo{booktitle}{\emph{FOCS}}.
\newblock


\bibitem[Azabou et~al\mbox{.}(2023)]%
        {addnode}
\bibfield{author}{\bibinfo{person}{Mehdi Azabou}, \bibinfo{person}{Venkataramana Ganesh}, \bibinfo{person}{Shantanu Thakoor}, \bibinfo{person}{Chi{-}Heng Lin}, \bibinfo{person}{Lakshmi Sathidevi}, \bibinfo{person}{Ran Liu}, \bibinfo{person}{Michal Valko}, \bibinfo{person}{Petar Velickovic}, {and} \bibinfo{person}{Eva~L. Dyer}.} \bibinfo{year}{2023}\natexlab{}.
\newblock \showarticletitle{Half-Hop: {A} graph upsampling approach for slowing down message passing}. In \bibinfo{booktitle}{\emph{ICML}}.
\newblock


\bibitem[Bojchevski et~al\mbox{.}(2020)]%
        {pprgo}
\bibfield{author}{\bibinfo{person}{Aleksandar Bojchevski}, \bibinfo{person}{Johannes Gasteiger}, \bibinfo{person}{Bryan Perozzi}, \bibinfo{person}{Amol Kapoor}, \bibinfo{person}{Martin Blais}, \bibinfo{person}{Benedek R{\'o}zemberczki}, \bibinfo{person}{Michal Lukasik}, {and} \bibinfo{person}{Stephan G{\"u}nnemann}.} \bibinfo{year}{2020}\natexlab{}.
\newblock \showarticletitle{Scaling graph neural networks with approximate pagerank}. In \bibinfo{booktitle}{\emph{KDD}}.
\newblock


\bibitem[Chen et~al\mbox{.}(2020a)]%
        {AdaEdge}
\bibfield{author}{\bibinfo{person}{Deli Chen}, \bibinfo{person}{Yankai Lin}, \bibinfo{person}{Wei Li}, \bibinfo{person}{Peng Li}, \bibinfo{person}{Jie Zhou}, {and} \bibinfo{person}{Xu Sun}.} \bibinfo{year}{2020}\natexlab{a}.
\newblock \showarticletitle{Measuring and Relieving the Over-Smoothing Problem for Graph Neural Networks from the Topological View}. In \bibinfo{booktitle}{\emph{The Thirty-Fourth {AAAI} Conference on Artificial Intelligence, {AAAI} 2020, The Thirty-Second Innovative Applications of Artificial Intelligence Conference, {IAAI} 2020, The Tenth {AAAI} Symposium on Educational Advances in Artificial Intelligence, {EAAI} 2020, New York, NY, USA, February 7-12, 2020}}.
\newblock


\bibitem[Chen et~al\mbox{.}(2020b)]%
        {gbp}
\bibfield{author}{\bibinfo{person}{Ming Chen}, \bibinfo{person}{Zhewei Wei}, \bibinfo{person}{Bolin Ding}, \bibinfo{person}{Yaliang Li}, \bibinfo{person}{Ye Yuan}, \bibinfo{person}{Xiaoyong Du}, {and} \bibinfo{person}{Ji-Rong Wen}.} \bibinfo{year}{2020}\natexlab{b}.
\newblock \showarticletitle{Scalable graph neural networks via bidirectional propagation}.
\newblock \bibinfo{journal}{\emph{Advances in neural information processing systems}}  \bibinfo{volume}{33} (\bibinfo{year}{2020}), \bibinfo{pages}{14556--14566}.
\newblock


\bibitem[Chen et~al\mbox{.}(2020c)]%
        {sdg}
\bibfield{author}{\bibinfo{person}{Ming Chen}, \bibinfo{person}{Zhewei Wei}, \bibinfo{person}{Zengfeng Huang}, \bibinfo{person}{Bolin Ding}, {and} \bibinfo{person}{Yaliang Li}.} \bibinfo{year}{2020}\natexlab{c}.
\newblock \showarticletitle{Simple and Deep Graph Convolutional Networks}. In \bibinfo{booktitle}{\emph{Proceedings of the 37th International Conference on Machine Learning, {ICML} 2020, 13-18 July 2020, Virtual Event}}.
\newblock


\bibitem[Chen et~al\mbox{.}(2020d)]%
        {IDGL}
\bibfield{author}{\bibinfo{person}{Yu Chen}, \bibinfo{person}{Lingfei Wu}, {and} \bibinfo{person}{Mohammed~J. Zaki}.} \bibinfo{year}{2020}\natexlab{d}.
\newblock \showarticletitle{Iterative Deep Graph Learning for Graph Neural Networks: Better and Robust Node Embeddings}. In \bibinfo{booktitle}{\emph{Advances in Neural Information Processing Systems 33: Annual Conference on Neural Information Processing Systems 2020, NeurIPS 2020, December 6-12, 2020, virtual}}.
\newblock


\bibitem[Ding et~al\mbox{.}(2023)]%
        {s3cl}
\bibfield{author}{\bibinfo{person}{Kaize Ding}, \bibinfo{person}{Yancheng Wang}, \bibinfo{person}{Yingzhen Yang}, {and} \bibinfo{person}{Huan Liu}.} \bibinfo{year}{2023}\natexlab{}.
\newblock \showarticletitle{Eliciting Structural and Semantic Global Knowledge in Unsupervised Graph Contrastive Learning}. In \bibinfo{booktitle}{\emph{Thirty-Seventh {AAAI} Conference on Artificial Intelligence, {AAAI} 2023, Thirty-Fifth Conference on Innovative Applications of Artificial Intelligence, {IAAI} 2023, Thirteenth Symposium on Educational Advances in Artificial Intelligence, {EAAI} 2023, Washington, DC, USA, February 7-14, 2023}}.
\newblock


\bibitem[Ding et~al\mbox{.}(2022a)]%
        {assu1}
\bibfield{author}{\bibinfo{person}{Kaize Ding}, \bibinfo{person}{Zhe Xu}, \bibinfo{person}{Hanghang Tong}, {and} \bibinfo{person}{Huan Liu}.} \bibinfo{year}{2022}\natexlab{a}.
\newblock \showarticletitle{Data Augmentation for Deep Graph Learning: {A} Survey}.
\newblock \bibinfo{journal}{\emph{{SIGKDD} Explor.}} \bibinfo{volume}{24}, \bibinfo{number}{2} (\bibinfo{year}{2022}).
\newblock


\bibitem[Ding et~al\mbox{.}(2022b)]%
        {assu2}
\bibfield{author}{\bibinfo{person}{Kaize Ding}, \bibinfo{person}{Chuxu Zhang}, \bibinfo{person}{Jie Tang}, \bibinfo{person}{Nitesh~V. Chawla}, {and} \bibinfo{person}{Huan Liu}.} \bibinfo{year}{2022}\natexlab{b}.
\newblock \showarticletitle{Toward Graph Minimally-Supervised Learning}. In \bibinfo{booktitle}{\emph{{KDD} '22: The 28th {ACM} {SIGKDD} Conference on Knowledge Discovery and Data Mining, Washington, DC, USA, August 14 - 18, 2022}}.
\newblock


\bibitem[Fang et~al\mbox{.}(2023)]%
        {DropMessage}
\bibfield{author}{\bibinfo{person}{Taoran Fang}, \bibinfo{person}{Zhiqing Xiao}, \bibinfo{person}{Chunping Wang}, \bibinfo{person}{Jiarong Xu}, \bibinfo{person}{Xuan Yang}, {and} \bibinfo{person}{Yang Yang}.} \bibinfo{year}{2023}\natexlab{}.
\newblock \showarticletitle{DropMessage: Unifying Random Dropping for Graph Neural Networks}. In \bibinfo{booktitle}{\emph{Thirty-Seventh {AAAI} Conference on Artificial Intelligence, {AAAI} 2023, Thirty-Fifth Conference on Innovative Applications of Artificial Intelligence, {IAAI} 2023, Thirteenth Symposium on Educational Advances in Artificial Intelligence, {EAAI} 2023, Washington, DC, USA, February 7-14, 2023}}.
\newblock


\bibitem[Feng et~al\mbox{.}(2022)]%
        {GRAND+}
\bibfield{author}{\bibinfo{person}{Wenzheng Feng}, \bibinfo{person}{Yuxiao Dong}, \bibinfo{person}{Tinglin Huang}, \bibinfo{person}{Ziqi Yin}, \bibinfo{person}{Xu Cheng}, \bibinfo{person}{Evgeny Kharlamov}, {and} \bibinfo{person}{Jie Tang}.} \bibinfo{year}{2022}\natexlab{}.
\newblock \showarticletitle{{GRAND+:} Scalable Graph Random Neural Networks}. In \bibinfo{booktitle}{\emph{{WWW} '22: The {ACM} Web Conference 2022, Virtual Event, Lyon, France, April 25 - 29, 2022}}.
\newblock


\bibitem[Feng et~al\mbox{.}(2020)]%
        {grand}
\bibfield{author}{\bibinfo{person}{Wenzheng Feng}, \bibinfo{person}{Jie Zhang}, \bibinfo{person}{Yuxiao Dong}, \bibinfo{person}{Yu Han}, \bibinfo{person}{Huanbo Luan}, \bibinfo{person}{Qian Xu}, \bibinfo{person}{Qiang Yang}, \bibinfo{person}{Evgeny Kharlamov}, {and} \bibinfo{person}{Jie Tang}.} \bibinfo{year}{2020}\natexlab{}.
\newblock \showarticletitle{Graph Random Neural Networks for Semi-Supervised Learning on Graphs}. In \bibinfo{booktitle}{\emph{NeurIPS}}.
\newblock


\bibitem[Franceschi et~al\mbox{.}(2019)]%
        {learnsturct2}
\bibfield{author}{\bibinfo{person}{Luca Franceschi}, \bibinfo{person}{Mathias Niepert}, \bibinfo{person}{Massimiliano Pontil}, {and} \bibinfo{person}{Xiao He}.} \bibinfo{year}{2019}\natexlab{}.
\newblock \bibinfo{title}{Learning Discrete Structures for Graph Neural Networks}.
\newblock


\bibitem[Guo and Mao(2021)]%
        {gmixup2}
\bibfield{author}{\bibinfo{person}{Hongyu Guo} {and} \bibinfo{person}{Yongyi Mao}.} \bibinfo{year}{2021}\natexlab{}.
\newblock \showarticletitle{Intrusion-Free Graph Mixup}.
\newblock \bibinfo{journal}{\emph{CoRR}}  \bibinfo{volume}{abs/2110.09344} (\bibinfo{year}{2021}).
\newblock


\bibitem[Hamilton et~al\mbox{.}(2017)]%
        {graphsage}
\bibfield{author}{\bibinfo{person}{Will Hamilton}, \bibinfo{person}{Zhitao Ying}, {and} \bibinfo{person}{Jure Leskovec}.} \bibinfo{year}{2017}\natexlab{}.
\newblock \showarticletitle{Inductive representation learning on large graphs}.
\newblock \bibinfo{journal}{\emph{Advances in neural information processing systems}}  \bibinfo{volume}{30} (\bibinfo{year}{2017}).
\newblock


\bibitem[Han et~al\mbox{.}(2022)]%
        {gmixup1}
\bibfield{author}{\bibinfo{person}{Xiaotian Han}, \bibinfo{person}{Zhimeng Jiang}, \bibinfo{person}{Ninghao Liu}, {and} \bibinfo{person}{Xia Hu}.} \bibinfo{year}{2022}\natexlab{}.
\newblock \bibinfo{title}{G-Mixup: Graph Data Augmentation for Graph Classification}.
\newblock


\bibitem[He et~al\mbox{.}(2020)]%
        {lightgcn}
\bibfield{author}{\bibinfo{person}{Xiangnan He}, \bibinfo{person}{Kuan Deng}, \bibinfo{person}{Xiang Wang}, \bibinfo{person}{Yan Li}, \bibinfo{person}{Yong{-}Dong Zhang}, {and} \bibinfo{person}{Meng Wang}.} \bibinfo{year}{2020}\natexlab{}.
\newblock \showarticletitle{LightGCN: Simplifying and Powering Graph Convolution Network for Recommendation}. In \bibinfo{booktitle}{\emph{SIGIR}}. \bibinfo{publisher}{{ACM}}, \bibinfo{pages}{639--648}.
\newblock


\bibitem[Holland et~al\mbox{.}(1983)]%
        {SBM}
\bibfield{author}{\bibinfo{person}{Paul~W Holland}, \bibinfo{person}{Kathryn~Blackmond Laskey}, {and} \bibinfo{person}{Samuel Leinhardt}.} \bibinfo{year}{1983}\natexlab{}.
\newblock \showarticletitle{Stochastic blockmodels: First steps}.
\newblock \bibinfo{journal}{\emph{Social networks}} \bibinfo{volume}{5}, \bibinfo{number}{2} (\bibinfo{year}{1983}).
\newblock


\bibitem[Jiang et~al\mbox{.}(2019)]%
        {learnsturct1}
\bibfield{author}{\bibinfo{person}{Bo Jiang}, \bibinfo{person}{Ziyan Zhang}, \bibinfo{person}{Doudou Lin}, \bibinfo{person}{Jin Tang}, {and} \bibinfo{person}{Bin Luo}.} \bibinfo{year}{2019}\natexlab{}.
\newblock \showarticletitle{Semi-Supervised Learning With Graph Learning-Convolutional Networks}. In \bibinfo{booktitle}{\emph{2019 IEEE/CVF Conference on Computer Vision and Pattern Recognition (CVPR)}}.
\newblock


\bibitem[Jin et~al\mbox{.}(2021)]%
        {simpgnn}
\bibfield{author}{\bibinfo{person}{Wei Jin}, \bibinfo{person}{Tyler Derr}, \bibinfo{person}{Yiqi Wang}, \bibinfo{person}{Yao Ma}, \bibinfo{person}{Zitao Liu}, {and} \bibinfo{person}{Jiliang Tang}.} \bibinfo{year}{2021}\natexlab{}.
\newblock \showarticletitle{Node Similarity Preserving Graph Convolutional Networks}. In \bibinfo{booktitle}{\emph{WSDM}}.
\newblock


\bibitem[Keetha et~al\mbox{.}(2022)]%
        {cvpr}
\bibfield{author}{\bibinfo{person}{Nikhil~Varma Keetha}, \bibinfo{person}{Chen Wang}, \bibinfo{person}{Yuheng Qiu}, \bibinfo{person}{Kuan Xu}, {and} \bibinfo{person}{Sebastian~A. Scherer}.} \bibinfo{year}{2022}\natexlab{}.
\newblock \showarticletitle{AirObject: {A} Temporally Evolving Graph Embedding for Object Identification}. In \bibinfo{booktitle}{\emph{CVPR}}.
\newblock


\bibitem[Kipf and Welling(2016)]%
        {GAE}
\bibfield{author}{\bibinfo{person}{Thomas~N. Kipf} {and} \bibinfo{person}{Max Welling}.} \bibinfo{year}{2016}\natexlab{}.
\newblock \showarticletitle{Variational Graph Auto-Encoders}.
\newblock \bibinfo{journal}{\emph{CoRR}}  \bibinfo{volume}{abs/1611.07308} (\bibinfo{year}{2016}).
\newblock


\bibitem[Kipf and Welling(2017)]%
        {gcn}
\bibfield{author}{\bibinfo{person}{Thomas~N. Kipf} {and} \bibinfo{person}{Max Welling}.} \bibinfo{year}{2017}\natexlab{}.
\newblock \showarticletitle{Semi-Supervised Classification with Graph Convolutional Networks}. In \bibinfo{booktitle}{\emph{ICLR}}.
\newblock


\bibitem[Klicpera et~al\mbox{.}(2019a)]%
        {ppnp}
\bibfield{author}{\bibinfo{person}{Johannes Klicpera}, \bibinfo{person}{Aleksandar Bojchevski}, {and} \bibinfo{person}{Stephan G{\"{u}}nnemann}.} \bibinfo{year}{2019}\natexlab{a}.
\newblock \showarticletitle{Predict then Propagate: Graph Neural Networks meet Personalized PageRank}. In \bibinfo{booktitle}{\emph{ICLR}}.
\newblock


\bibitem[Klicpera et~al\mbox{.}(2019b)]%
        {DBLP:ppr}
\bibfield{author}{\bibinfo{person}{Johannes Klicpera}, \bibinfo{person}{Aleksandar Bojchevski}, {and} \bibinfo{person}{Stephan G{\"{u}}nnemann}.} \bibinfo{year}{2019}\natexlab{b}.
\newblock \showarticletitle{Predict then Propagate: Graph Neural Networks meet Personalized PageRank}. In \bibinfo{booktitle}{\emph{7th International Conference on Learning Representations, {ICLR} 2019, New Orleans, LA, USA, May 6-9, 2019}}.
\newblock


\bibitem[Li et~al\mbox{.}(2022)]%
        {glognn}
\bibfield{author}{\bibinfo{person}{Xiang Li}, \bibinfo{person}{Renyu Zhu}, \bibinfo{person}{Yao Cheng}, \bibinfo{person}{Caihua Shan}, \bibinfo{person}{Siqiang Luo}, \bibinfo{person}{Dongsheng Li}, {and} \bibinfo{person}{Weining Qian}.} \bibinfo{year}{2022}\natexlab{}.
\newblock \showarticletitle{Finding Global Homophily in Graph Neural Networks When Meeting Heterophily}. In \bibinfo{booktitle}{\emph{International Conference on Machine Learning, {ICML} 2022, 17-23 July 2022, Baltimore, Maryland, {USA}}}.
\newblock


\bibitem[Liao et~al\mbox{.}(2024)]%
        {DBLP:journals/entropy/LiaoLHL24}
\bibfield{author}{\bibinfo{person}{Ziyu Liao}, \bibinfo{person}{Tao Liu}, \bibinfo{person}{Yue He}, {and} \bibinfo{person}{Longlong Lin}.} \bibinfo{year}{2024}\natexlab{}.
\newblock \showarticletitle{Effective Temporal Graph Learning via Personalized PageRank}.
\newblock \bibinfo{journal}{\emph{Entropy}} \bibinfo{volume}{26}, \bibinfo{number}{7} (\bibinfo{year}{2024}), \bibinfo{pages}{588}.
\newblock


\bibitem[Lin et~al\mbox{.}(2024)]%
        {DBLP:conf/cikm/LinYWWZ0024}
\bibfield{author}{\bibinfo{person}{Longlong Lin}, \bibinfo{person}{Yunfeng Yu}, \bibinfo{person}{Zihao Wang}, \bibinfo{person}{Zeli Wang}, \bibinfo{person}{Yuying Zhao}, \bibinfo{person}{Jin Zhao}, {and} \bibinfo{person}{Tao Jia}.} \bibinfo{year}{2024}\natexlab{}.
\newblock \showarticletitle{{PSNE:} Efficient Spectral Sparsification Algorithms for Scaling Network Embedding}. In \bibinfo{booktitle}{\emph{CIKM}}. \bibinfo{pages}{1420--1429}.
\newblock


\bibitem[Liu et~al\mbox{.}(2023)]%
        {ddpt}
\bibfield{author}{\bibinfo{person}{Yixin Liu}, \bibinfo{person}{Kaize Ding}, \bibinfo{person}{Jianling Wang}, \bibinfo{person}{Vincent C.~S. Lee}, \bibinfo{person}{Huan Liu}, {and} \bibinfo{person}{Shirui Pan}.} \bibinfo{year}{2023}\natexlab{}.
\newblock \showarticletitle{Learning Strong Graph Neural Networks with Weak Information}. In \bibinfo{booktitle}{\emph{KDD 2023, Long Beach, CA, USA, August 6-10, 2023}}.
\newblock


\bibitem[Luan et~al\mbox{.}(2022)]%
        {luansitao}
\bibfield{author}{\bibinfo{person}{Sitao Luan}, \bibinfo{person}{Chenqing Hua}, \bibinfo{person}{Qincheng Lu}, \bibinfo{person}{Jiaqi Zhu}, \bibinfo{person}{Mingde Zhao}, \bibinfo{person}{Shuyuan Zhang}, \bibinfo{person}{Xiao{-}Wen Chang}, {and} \bibinfo{person}{Doina Precup}.} \bibinfo{year}{2022}\natexlab{}.
\newblock \showarticletitle{Revisiting Heterophily For Graph Neural Networks}. In \bibinfo{booktitle}{\emph{Advances in Neural Information Processing Systems 35:Annual Conference on Neural Information Processing Systems 2022, NeurIPS 2022, New Orleans, LA, USA, November 28 - December 9, 2022}}.
\newblock


\bibitem[Luo et~al\mbox{.}(2021)]%
        {learndrop}
\bibfield{author}{\bibinfo{person}{Dongsheng Luo}, \bibinfo{person}{Wei Cheng}, \bibinfo{person}{Wenchao Yu}, \bibinfo{person}{Bo Zong}, \bibinfo{person}{Jingchao Ni}, \bibinfo{person}{Haifeng Chen}, {and} \bibinfo{person}{Xiang Zhang}.} \bibinfo{year}{2021}\natexlab{}.
\newblock \showarticletitle{Learning to Drop: Robust Graph Neural Network via Topological Denoising}. In \bibinfo{booktitle}{\emph{{WSDM}}}.
\newblock


\bibitem[Meng et~al\mbox{.}(2024)]%
        {DBLP:journals/pvldb/MengLLLW24}
\bibfield{author}{\bibinfo{person}{Yuchen Meng}, \bibinfo{person}{Ronghua Li}, \bibinfo{person}{Longlong Lin}, \bibinfo{person}{Xunkai Li}, {and} \bibinfo{person}{Guoren Wang}.} \bibinfo{year}{2024}\natexlab{}.
\newblock \showarticletitle{Topology-preserving Graph Coarsening: An Elementary Collapse-based Approach}.
\newblock \bibinfo{journal}{\emph{Proc. {VLDB} Endow.}} \bibinfo{volume}{17}, \bibinfo{number}{13} (\bibinfo{year}{2024}), \bibinfo{pages}{4760--4772}.
\newblock


\bibitem[Park et~al\mbox{.}(2021)]%
        {MH-Aug}
\bibfield{author}{\bibinfo{person}{Hyeon{-}Jin Park}, \bibinfo{person}{Seunghun Lee}, \bibinfo{person}{Sihyeon Kim}, \bibinfo{person}{Jinyoung Park}, \bibinfo{person}{Jisu Jeong}, \bibinfo{person}{Kyung{-}Min Kim}, \bibinfo{person}{Jung{-}Woo Ha}, {and} \bibinfo{person}{Hyunwoo~J. Kim}.} \bibinfo{year}{2021}\natexlab{}.
\newblock \showarticletitle{Metropolis-Hastings Data Augmentation for Graph Neural Networks}. In \bibinfo{booktitle}{\emph{Advances in Neural Information Processing Systems 34: Annual Conference on Neural Information Processing Systems 2021, NeurIPS 2021, December 6-14, 2021, virtual}}.
\newblock


\bibitem[Platonov et~al\mbox{.}(2023)]%
        {heterograph}
\bibfield{author}{\bibinfo{person}{Oleg Platonov}, \bibinfo{person}{Denis Kuznedelev}, \bibinfo{person}{Michael Diskin}, \bibinfo{person}{Artem Babenko}, {and} \bibinfo{person}{Liudmila Prokhorenkova}.} \bibinfo{year}{2023}\natexlab{}.
\newblock \showarticletitle{A critical look at the evaluation of GNNs under heterophily: Are we really making progress?}
\newblock  (\bibinfo{year}{2023}).
\newblock


\bibitem[Qiu et~al\mbox{.}(2018)]%
        {social1}
\bibfield{author}{\bibinfo{person}{Jiezhong Qiu}, \bibinfo{person}{Jian Tang}, \bibinfo{person}{Hao Ma}, \bibinfo{person}{Yuxiao Dong}, \bibinfo{person}{Kuansan Wang}, {and} \bibinfo{person}{Jie Tang}.} \bibinfo{year}{2018}\natexlab{}.
\newblock \showarticletitle{DeepInf: Social Influence Prediction with Deep Learning}. In \bibinfo{booktitle}{\emph{SIGKDD}}.
\newblock


\bibitem[Rong et~al\mbox{.}(2020)]%
        {sturct4:}
\bibfield{author}{\bibinfo{person}{Yu Rong}, \bibinfo{person}{Wenbing Huang}, \bibinfo{person}{Tingyang Xu}, {and} \bibinfo{person}{Junzhou Huang}.} \bibinfo{year}{2020}\natexlab{}.
\newblock \showarticletitle{DropEdge: Towards Deep Graph Convolutional Networks on Node Classification}. In \bibinfo{booktitle}{\emph{International Conference on Learning Representations}}.
\newblock


\bibitem[Seibold et~al\mbox{.}(2022)]%
        {pl}
\bibfield{author}{\bibinfo{person}{Constantin~Marc Seibold}, \bibinfo{person}{Simon Rei{\ss}}, \bibinfo{person}{Jens Kleesiek}, {and} \bibinfo{person}{Rainer Stiefelhagen}.} \bibinfo{year}{2022}\natexlab{}.
\newblock \showarticletitle{Reference-Guided Pseudo-Label Generation for Medical Semantic Segmentation}. In \bibinfo{booktitle}{\emph{Thirty-Sixth {AAAI} Conference on Artificial Intelligence, {AAAI} 2022, Thirty-Fourth Conference on Innovative Applications of Artificial Intelligence, {IAAI} 2022, The Twelveth Symposium on Educational Advances in Artificial Intelligence, {EAAI} 2022 Virtual Event, February 22 - March 1, 2022}}.
\newblock


\bibitem[Sen et~al\mbox{.}(2008)]%
        {cora}
\bibfield{author}{\bibinfo{person}{Prithviraj Sen}, \bibinfo{person}{Galileo Namata}, \bibinfo{person}{Mustafa Bilgic}, \bibinfo{person}{Lise Getoor}, \bibinfo{person}{Brian Gallagher}, {and} \bibinfo{person}{Tina Eliassi{-}Rad}.} \bibinfo{year}{2008}\natexlab{}.
\newblock \showarticletitle{Collective Classification in Network Data}.
\newblock \bibinfo{journal}{\emph{{AI} Mag.}} \bibinfo{volume}{29}, \bibinfo{number}{3} (\bibinfo{year}{2008}).
\newblock


\bibitem[Shang et~al\mbox{.}(2021)]%
        {GTS}
\bibfield{author}{\bibinfo{person}{Chao Shang}, \bibinfo{person}{Jie Chen}, {and} \bibinfo{person}{Jinbo Bi}.} \bibinfo{year}{2021}\natexlab{}.
\newblock \showarticletitle{Discrete Graph Structure Learning for Forecasting Multiple Time Series}. In \bibinfo{booktitle}{\emph{9th International Conference on Learning Representations, {ICLR} 2021, Virtual Event, Austria, May 3-7, 2021}}.
\newblock


\bibitem[Shchur et~al\mbox{.}(2018)]%
        {coauthor}
\bibfield{author}{\bibinfo{person}{Oleksandr Shchur}, \bibinfo{person}{Maximilian Mumme}, \bibinfo{person}{Aleksandar Bojchevski}, {and} \bibinfo{person}{Stephan G{\"{u}}nnemann}.} \bibinfo{year}{2018}\natexlab{}.
\newblock \showarticletitle{Pitfalls of Graph Neural Network Evaluation}.
\newblock \bibinfo{journal}{\emph{CoRR}} (\bibinfo{year}{2018}).
\newblock


\bibitem[Thakoor et~al\mbox{.}(2022)]%
        {bgrl}
\bibfield{author}{\bibinfo{person}{Shantanu Thakoor}, \bibinfo{person}{Corentin Tallec}, \bibinfo{person}{Mohammad~Gheshlaghi Azar}, \bibinfo{person}{Mehdi Azabou}, \bibinfo{person}{Eva~L. Dyer}, \bibinfo{person}{R{\'{e}}mi Munos}, \bibinfo{person}{Petar Velickovic}, {and} \bibinfo{person}{Michal Valko}.} \bibinfo{year}{2022}\natexlab{}.
\newblock \showarticletitle{Large-Scale Representation Learning on Graphs via Bootstrapping}. In \bibinfo{booktitle}{\emph{The Tenth International Conference on Learning Representations, {ICLR} 2022, Virtual Event, April 25-29, 2022}}.
\newblock


\bibitem[Veličković et~al\mbox{.}(2018)]%
        {gat}
\bibfield{author}{\bibinfo{person}{Petar Veličković}, \bibinfo{person}{Guillem Cucurull}, \bibinfo{person}{Arantxa Casanova}, \bibinfo{person}{Adriana Romero}, \bibinfo{person}{Pietro Liò}, {and} \bibinfo{person}{Yoshua Bengio}.} \bibinfo{year}{2018}\natexlab{}.
\newblock \showarticletitle{Graph Attention Networks}. In \bibinfo{booktitle}{\emph{International Conference on Learning Representations}}.
\newblock


\bibitem[Verma et~al\mbox{.}(2021)]%
        {graphmix}
\bibfield{author}{\bibinfo{person}{Vikas Verma}, \bibinfo{person}{Meng Qu}, \bibinfo{person}{Kenji Kawaguchi}, \bibinfo{person}{Alex Lamb}, \bibinfo{person}{Yoshua Bengio}, \bibinfo{person}{Juho Kannala}, {and} \bibinfo{person}{Jian Tang}.} \bibinfo{year}{2021}\natexlab{}.
\newblock \showarticletitle{GraphMix: Improved Training of GNNs for Semi-Supervised Learning}. In \bibinfo{booktitle}{\emph{Thirty-Fifth {AAAI} Conference on Artificial Intelligence, {AAAI} 2021, Thirty-Third Conference on Innovative Applications of Artificial Intelligence, {IAAI} 2021, The Eleventh Symposium on Educational Advances in Artificial Intelligence, {EAAI} 2021, Virtual Event, February 2-9, 2021}}.
\newblock


\bibitem[Wang et~al\mbox{.}(2021a)]%
        {GEN}
\bibfield{author}{\bibinfo{person}{Ruijia Wang}, \bibinfo{person}{Shuai Mou}, \bibinfo{person}{Xiao Wang}, \bibinfo{person}{Wanpeng Xiao}, \bibinfo{person}{Qi Ju}, \bibinfo{person}{Chuan Shi}, {and} \bibinfo{person}{Xing Xie}.} \bibinfo{year}{2021}\natexlab{a}.
\newblock \showarticletitle{Graph Structure Estimation Neural Networks}. In \bibinfo{booktitle}{\emph{{WWW} '21: The Web Conference 2021, Virtual Event / Ljubljana, Slovenia, April 19-23, 2021}}.
\newblock


\bibitem[Wang et~al\mbox{.}(2017)]%
        {social2}
\bibfield{author}{\bibinfo{person}{Suhang Wang}, \bibinfo{person}{Jiliang Tang}, \bibinfo{person}{Charu~C. Aggarwal}, \bibinfo{person}{Yi Chang}, {and} \bibinfo{person}{Huan Liu}.} \bibinfo{year}{2017}\natexlab{}.
\newblock \showarticletitle{Signed Network Embedding in Social Media}. In \bibinfo{booktitle}{\emph{SIAM}}.
\newblock


\bibitem[Wang and Isola(2020)]%
        {clmit}
\bibfield{author}{\bibinfo{person}{Tongzhou Wang} {and} \bibinfo{person}{Phillip Isola}.} \bibinfo{year}{2020}\natexlab{}.
\newblock \showarticletitle{Understanding Contrastive Representation Learning through Alignment and Uniformity on the Hypersphere}. In \bibinfo{booktitle}{\emph{Proceedings of the 37th International Conference on Machine Learning, {ICML} 2020, 13-18 July 2020, Virtual Event}}.
\newblock


\bibitem[Wang et~al\mbox{.}(2021b)]%
        {mixup}
\bibfield{author}{\bibinfo{person}{Yiwei Wang}, \bibinfo{person}{Wei Wang}, \bibinfo{person}{Yuxuan Liang}, \bibinfo{person}{Yujun Cai}, {and} \bibinfo{person}{Bryan Hooi}.} \bibinfo{year}{2021}\natexlab{b}.
\newblock \showarticletitle{Mixup for Node and Graph Classification}. In \bibinfo{booktitle}{\emph{{WWW} '21: The Web Conference 2021, Virtual Event / Ljubljana, Slovenia, April 19-23, 2021}}.
\newblock


\bibitem[Wang et~al\mbox{.}(2024a)]%
        {DBLP:conf/mir/WangLXL024}
\bibfield{author}{\bibinfo{person}{Zeli Wang}, \bibinfo{person}{Jian Li}, \bibinfo{person}{Shuyin Xia}, \bibinfo{person}{Longlong Lin}, {and} \bibinfo{person}{Guoyin Wang}.} \bibinfo{year}{2024}\natexlab{a}.
\newblock \showarticletitle{Text Adversarial Defense via Granular-Ball Sample Enhancement}. In \bibinfo{booktitle}{\emph{ICMR}}. \bibinfo{pages}{348--356}.
\newblock


\bibitem[Wang et~al\mbox{.}(2024b)]%
        {DBLP:conf/mir/WangZXLW24}
\bibfield{author}{\bibinfo{person}{Zeli Wang}, \bibinfo{person}{Tuo Zhang}, \bibinfo{person}{Shuyin Xia}, \bibinfo{person}{Longlong Lin}, {and} \bibinfo{person}{Guoyin Wang}.} \bibinfo{year}{2024}\natexlab{b}.
\newblock \showarticletitle{GB\({}_{\mbox{RAIN}}\): Combating Textual Label Noise by Granular-ball based Robust Training}. In \bibinfo{booktitle}{\emph{ICMR}}. \bibinfo{pages}{357--365}.
\newblock


\bibitem[Wu et~al\mbox{.}(2019)]%
        {sgc}
\bibfield{author}{\bibinfo{person}{Felix Wu}, \bibinfo{person}{Amauri Souza}, \bibinfo{person}{Tianyi Zhang}, \bibinfo{person}{Christopher Fifty}, \bibinfo{person}{Tao Yu}, {and} \bibinfo{person}{Kilian Weinberger}.} \bibinfo{year}{2019}\natexlab{}.
\newblock \showarticletitle{Simplifying Graph Convolutional Networks}. In \bibinfo{booktitle}{\emph{Proceedings of the 36th International Conference on Machine Learning, {ICML} 2019, 9-15 June 2019, Long Beach, California, {USA}}}.
\newblock


\bibitem[Xia et~al\mbox{.}(2022)]%
        {progcl}
\bibfield{author}{\bibinfo{person}{Jun Xia}, \bibinfo{person}{Lirong Wu}, \bibinfo{person}{Ge Wang}, \bibinfo{person}{Jintao Chen}, {and} \bibinfo{person}{Stan~Z. Li}.} \bibinfo{year}{2022}\natexlab{}.
\newblock \showarticletitle{ProGCL: Rethinking Hard Negative Mining in Graph Contrastive Learning}. In \bibinfo{booktitle}{\emph{International Conference on Machine Learning, {ICML} 2022, 17-23 July 2022, Baltimore, Maryland, {USA}}}.
\newblock


\bibitem[Xue et~al\mbox{.}(2024)]%
        {psagnn}
\bibfield{author}{\bibinfo{person}{Guotong Xue}, \bibinfo{person}{Ming Zhong}, \bibinfo{person}{Tieyun Qian}, {and} \bibinfo{person}{Jianxin Li}.} \bibinfo{year}{2024}\natexlab{}.
\newblock \showarticletitle{{PSA-GNN:} An augmented {GNN} framework with priori subgraph knowledge}.
\newblock \bibinfo{journal}{\emph{Neural Networks}}  \bibinfo{volume}{173} (\bibinfo{year}{2024}).
\newblock


\bibitem[Yang et~al\mbox{.}(2019)]%
        {TO-GCN}
\bibfield{author}{\bibinfo{person}{Liang Yang}, \bibinfo{person}{Zesheng Kang}, \bibinfo{person}{Xiaochun Cao}, \bibinfo{person}{Di Jin}, \bibinfo{person}{Bo Yang}, {and} \bibinfo{person}{Yuanfang Guo}.} \bibinfo{year}{2019}\natexlab{}.
\newblock \showarticletitle{Topology Optimization based Graph Convolutional Network}. In \bibinfo{booktitle}{\emph{Proceedings of the Twenty-Eighth International Joint Conference on Artificial Intelligence, {IJCAI} 2019, Macao, China, August 10-16, 2019}}.
\newblock


\bibitem[Ying et~al\mbox{.}(2018)]%
        {DBLP:conf/kdd/YingHCEHL18}
\bibfield{author}{\bibinfo{person}{Rex Ying}, \bibinfo{person}{Ruining He}, \bibinfo{person}{Kaifeng Chen}, \bibinfo{person}{Pong Eksombatchai}, \bibinfo{person}{William~L. Hamilton}, {and} \bibinfo{person}{Jure Leskovec}.} \bibinfo{year}{2018}\natexlab{}.
\newblock \showarticletitle{Graph Convolutional Neural Networks for Web-Scale Recommender Systems}. In \bibinfo{booktitle}{\emph{SIGKDD}}.
\newblock


\bibitem[Yu et~al\mbox{.}(2024)]%
        {DBLP:conf/mir/YuLLWOJ24}
\bibfield{author}{\bibinfo{person}{Yunfeng Yu}, \bibinfo{person}{Longlong Lin}, \bibinfo{person}{Qiyu Liu}, \bibinfo{person}{Zeli Wang}, \bibinfo{person}{Xi Ou}, {and} \bibinfo{person}{Tao Jia}.} \bibinfo{year}{2024}\natexlab{}.
\newblock \showarticletitle{{GSD-GNN:} Generalizable and Scalable Algorithms for Decoupled Graph Neural Networks}. In \bibinfo{booktitle}{\emph{ICMR}}. \bibinfo{pages}{64--72}.
\newblock


\bibitem[Zhao et~al\mbox{.}(2021b)]%
        {plabels}
\bibfield{author}{\bibinfo{person}{Han Zhao}, \bibinfo{person}{Xu Yang}, \bibinfo{person}{Zhenru Wang}, \bibinfo{person}{Erkun Yang}, {and} \bibinfo{person}{Cheng Deng}.} \bibinfo{year}{2021}\natexlab{b}.
\newblock \showarticletitle{Graph Debiased Contrastive Learning with Joint Representation Clustering}. In \bibinfo{booktitle}{\emph{Proceedings of the Thirtieth International Joint Conference on Artificial Intelligence, {IJCAI} 2021, Virtual Event / Montreal, Canada, 19-27 August 2021}}.
\newblock


\bibitem[Zhao et~al\mbox{.}(2019)]%
        {cvpr2}
\bibfield{author}{\bibinfo{person}{Long Zhao}, \bibinfo{person}{Xi Peng}, \bibinfo{person}{Yu Tian}, \bibinfo{person}{Mubbasir Kapadia}, {and} \bibinfo{person}{Dimitris~N. Metaxas}.} \bibinfo{year}{2019}\natexlab{}.
\newblock \showarticletitle{Semantic Graph Convolutional Networks for 3D Human Pose Regression}. In \bibinfo{booktitle}{\emph{CVPR}}.
\newblock


\bibitem[Zhao et~al\mbox{.}(2021a)]%
        {learnsturct3}
\bibfield{author}{\bibinfo{person}{Tong Zhao}, \bibinfo{person}{Yozen Liu}, \bibinfo{person}{Leonardo Neves}, \bibinfo{person}{Oliver Woodford}, \bibinfo{person}{Meng Jiang}, {and} \bibinfo{person}{Neil Shah}.} \bibinfo{year}{2021}\natexlab{a}.
\newblock \bibinfo{title}{Data Augmentation for Graph Neural Networks}.
\newblock


\bibitem[Zhao et~al\mbox{.}(2024)]%
        {geomix}
\bibfield{author}{\bibinfo{person}{Wentao Zhao}, \bibinfo{person}{Qitian Wu}, \bibinfo{person}{Chenxiao Yang}, {and} \bibinfo{person}{Junchi Yan}.} \bibinfo{year}{2024}\natexlab{}.
\newblock \showarticletitle{GeoMix: Towards Geometry-Aware Data Augmentation}. In \bibinfo{booktitle}{\emph{Proceedings of the 30th {ACM} {SIGKDD} Conference on Knowledge Discovery and Data Mining, {KDD} 2024, Barcelona, Spain, August 25-29, 2024}}.
\newblock


\bibitem[Zhu et~al\mbox{.}(2020)]%
        {GRACE}
\bibfield{author}{\bibinfo{person}{Yanqiao Zhu}, \bibinfo{person}{Yichen Xu}, \bibinfo{person}{Feng Yu}, \bibinfo{person}{Qiang Liu}, \bibinfo{person}{Shu Wu}, {and} \bibinfo{person}{Liang Wang}.} \bibinfo{year}{2020}\natexlab{}.
\newblock \showarticletitle{Deep Graph Contrastive Representation Learning}.
\newblock \bibinfo{journal}{\emph{CoRR}}  \bibinfo{volume}{abs/2006.04131} (\bibinfo{year}{2020}).
\newblock


\bibitem[Zhuang and Ma(2018)]%
        {dgc}
\bibfield{author}{\bibinfo{person}{Chenyi Zhuang} {and} \bibinfo{person}{Qiang Ma}.} \bibinfo{year}{2018}\natexlab{}.
\newblock \showarticletitle{Dual Graph Convolutional Networks for Graph-Based Semi-Supervised Classification}. In \bibinfo{booktitle}{\emph{Proceedings of the 2018 World Wide Web Conference on World Wide Web, {WWW} 2018, Lyon, France, April 23-27, 2018}}.
\newblock


\end{thebibliography}

\balance

\end{document}